\documentclass[lettersize,journal]{IEEEtran}
\usepackage{amsmath,amsfonts}
\usepackage{algorithmic}
\usepackage{algorithm}
\usepackage{array}
\usepackage[caption=false,font=footnotesize,labelfont=rm,textfont=rm]{subfig}
\usepackage{textcomp}
\usepackage{stfloats}
\usepackage{url}
\usepackage{verbatim}
\usepackage{graphicx}
\usepackage{cite}

\usepackage{algorithm}
\usepackage{algorithmic}

\usepackage{multirow}
\usepackage{amsmath}
\usepackage{amssymb}
\usepackage{pifont}
 \usepackage{booktabs}
\usepackage[table]{xcolor}
\begin{document}

\title{Synthetic-to-Real Translation for Class-Agnostic Motion Prediction}

\author{Yizheng Wu\textsuperscript{*}, 
         Hongwei Fan\textsuperscript{*},
        Kewei Wang,
         Ruibo Li,
         Xingyi Li,
        Xiao Song,
        Zhe Wang,
        Chenjing Ding,
        Dongliang Wang,
        Zhiguo Cao,
        Guosheng Lin
\IEEEcompsocitemizethanks{\IEEEcompsocthanksitem Yizheng Wu, Kewei Wang, and Xingyi Li are with the Key Laboratory
of Image Processing and Intelligent Control, Ministry of Education, School of Artificial Intelligence and Automation, Huazhong University of Science and Technology, Wuhan 430074, China, and also with Nanyang Technological University, Singapore 639798, Singapore.
\IEEEcompsocthanksitem Hongwei Fan, Xiao Song, Zhe Wang, Chenjing Ding, and Dongliang Wang are with the SenseAuto, Beijing 100080, China.
\IEEEcompsocthanksitem Zhiguo Cao is with the Key Laboratory of Image Processing and Intelligent Control, Ministry of Education, School of Artificial Intelligence and Automation, Huazhong University of Science and Technology, Wuhan 430074, China.
\IEEEcompsocthanksitem Ruibo Li, Guosheng Lin is with Nanyang Technological University, Singapore 639798, Singapore. 
\IEEEcompsocthanksitem Corresponding author: Guosheng Lin. Email: gslin@ntu.edu.sg
\IEEEcompsocthanksitem Yizheng Wu and Hongwei Fan contributed equally.
\IEEEcompsocthanksitem This paper has supplementary downloadable material available at http://ieeexplore.ieee.org., provided by the author. The material includes a video comparing our dataset with other real-world datasets. This material is 142MB in size.}}

\markboth{Accepted for Publication in IEEE Transactions on Multimedia}%
{Shell \MakeLowercase{\textit{et al.}}: A Sample Article Using IEEEtran.cls for IEEE Journals}

\IEEEpubid{0000--0000/00\$00.00~\copyright~2021 IEEE}

\maketitle

\begin{abstract}
Motion understanding is critical for ensuring safety and robustness in autonomous driving systems, driving increasing interest in motion prediction. A key challenge in this domain is the high cost associated with acquiring real-world motion labels. It is therefore ideal if we could transfer motion knowledge from synthetic data to real data. In this context, we explore the potential of synthetic-to-real translation for motion prediction (SRMP). However, the most used naive motion regression methods are notably sensitive to the synthetic-to-real domain shift, resulting in unreliable knowledge translation. To address this, we propose a novel approach integrating a motion knowledge translation framework with two key components: (1) objectness-aware motion prediction, which explicitly models the joint distribution of motion patterns and objectness priors to improve domain-invariant feature learning, and (2) objectness-aided motion enhancement, a motion label refinement mechanism that leverages learned objectness priors to filter motion noise. Furthermore, we present a physically-based pipeline for generating Motion4D, the first synthetic 4D LiDAR dataset tailored for SRMP research, addressing the lack of synthetic motion datasets. Experimental results demonstrate that our approach effectively bridges the domain gaps and yields superior performance on real scenes. Code and dataset will be made publicly available.

\end{abstract}    
\section{Introduction}
\label{sec:intro}

Motion prediction~\cite{chai2020multipath,liang2020pnpnet,motionnet,wang2022sti,dong2024sparse} aims to predict future motion displacements based on past observations from multimodal sensors, playing a crucial role in path planning and navigation for autonomous driving systems. To mitigate failures when dealing with unknown classes, many recent motion prediction approaches~\cite{mt,li2023weakly,wang2022sti} adopt a class-agnostic paradigm to directly regress cell-wise motions. These approaches transform the point clouds into bird's eye view (BEV) maps and represent motions as 2D displacements along the ground plane, achieving convincing performance. However, it is extremely challenging and costly to obtain motion labels on real data due to the sparse and non-uniform nature of point clouds~\cite{waymo-perception,caesar2020nuscenes}. In contrast, the synthetic 4D LiDAR sequences can directly provide accurate motion labels. The availability of labeled synthetic motion data thus prompts us to explore synthetic-to-real translation for motion prediction (SRMP), where the motion prediction model learns knowledge from synthetic data and transfers it to unlabeled real-world data. 

\begin{figure}[!t]
\centering
\subfloat[Baseline.]{\includegraphics[width=0.34\linewidth]{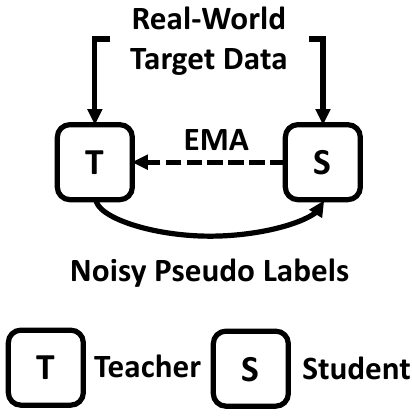}}
\subfloat[Ours (SR-Motion).]{\includegraphics[width=0.63\linewidth]{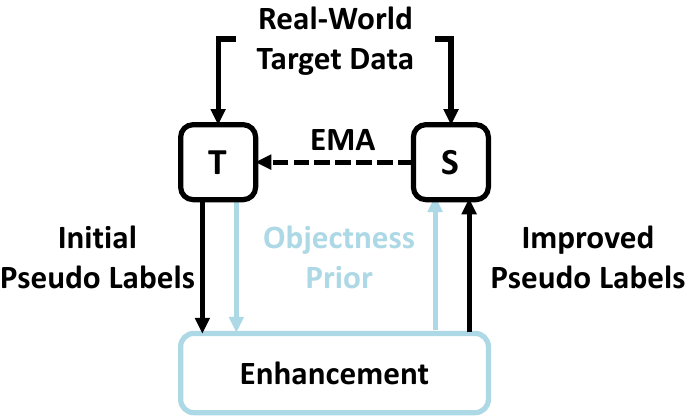}}\\
\subfloat[Visualization of motions.]{\includegraphics[width=0.9\linewidth]{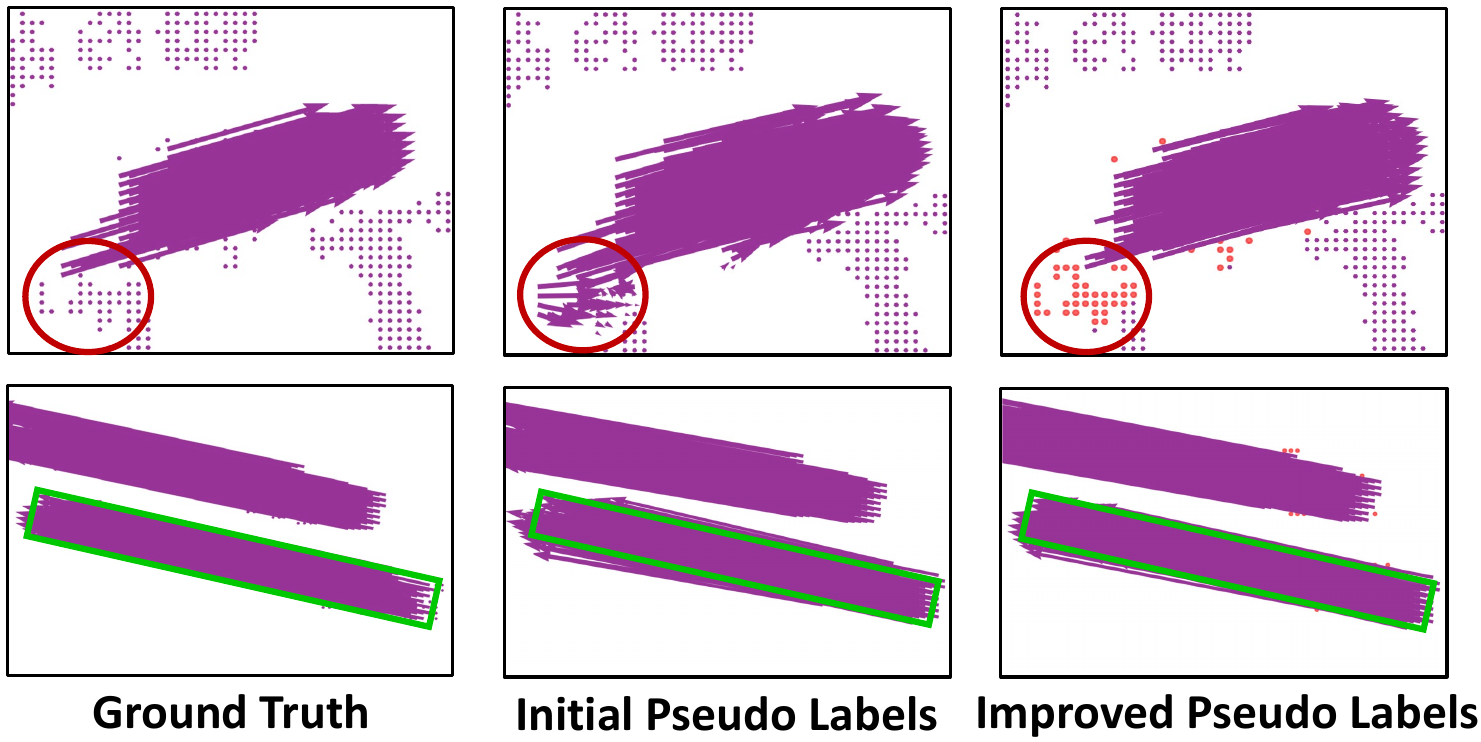}}
\caption{\textbf{Problem identification and resolution for SRMP.} (a) The naive regression of motions results in noisy pseudo labels. (b) SR-Motion addresses this issue by learning and leveraging objectness priors. (c) The purple arrows indicate cell-wise motions. Motion jitters (highlighted by red cycles) and object-level inconsistent predictions (stressed by green boxes) in initial pseudo labels hinder the stable knowledge translation. Orange cells are excluded noisy outliers. SR-Motion enhances initial pseudo labels as improved pseudo labels. }
\label{fig:synthetic-to-real}
\vspace{-0.7cm}
\end{figure}

\IEEEpubidadjcol 
While unsupervised domain adaptation has achieved notable success in tasks such as semantic segmentation and object detection~\cite{mt,yang2021st3d,yang2022st3d++,wang2022cross}, its extension to synthetic-to-real motion prediction remains underexplored. Motion prediction is characterized by a significant domain gap between synthetic and real-world data, stemming from discrepancies in motion distributions, occlusion patterns, and object compositions. This gap is particularly pronounced in the synthetic-to-real setting; consequently, a model trained solely on synthetic data suffers a severe performance degradation when directly deployed in the real world, exhibiting a $150.5\%$ increase in prediction error. A common approach to bridge this synthetic-to-real gap is the teacher-student framework~\cite{mt,yang2021st3d,yang2022st3d++,gta-sf,karim2023c,zhang2021prototypical,pan2024pseudo}, where a teacher model generates pseudo-labels on real data to supervise a student model. However, a naive application of this framework to motion prediction (Fig.~\ref{fig:synthetic-to-real}(a)) proves ineffective, further degrading performance with a $202.7\%$ error increase. We attribute this failure to noisy pseudo-labels generated by the teacher, which impede effective knowledge transfer. Specifically, we observe two dominant failure modes in the pseudo-labels (Fig.~\ref{fig:synthetic-to-real}(c)): (1) motion jitters, characterized by erratic and non-smooth motion vectors at the individual cell level, and (2) object-level inconsistencies, where cells belonging to the same physical object exhibit divergent or misaligned motion patterns. Prior work~\cite{motionnet} suggests that such artifacts are inherent to standard motion regression methods, which typically operate in a cell-wise, context-agnostic manner, rendering them highly vulnerable to domain shifts.

To overcome this, we propose SR-Motion, a novel framework that exploits objectness as a structural prior to mitigate the sensitivity of motion regression under domain shift. Inspired by MotionNet~\cite{motionnet}, which uses ground-truth object labels to regularize motion consistency within instances, we aim to leverage objectness priors to enhance pseudo label reliability. However, this is challenging in synthetic-to-real motion prediction, where real-world data lacks object annotations and raw motion vectors are too noisy for reliable clustering. To address this, SR-Motion introduces two key components: Objectness-aware motion prediction, in which the model jointly predicts motion vectors and objectness priors, specifically, the relative offset from each point to its object centroid, enabling the network to implicitly capture object-level structure during motion estimation; and Objectness-aided motion enhancement, which employs a dual-path consistency mechanism: only points where both the objectness and motion branches agree on cluster assignment are retained and spatially smoothed. This mutual validation between objectness and motion suppresses jitter and enforces coherent object-level motion, significantly improving the quality of pseudo labels and enabling stable knowledge transfer.

Furthermore, since the exploration into SRMP remains at an early stage, there is a lack of synthetic datasets specifically designed for SRMP. To address this, we develop a practical synthesis pipeline and construct a large-scale synthetic LiDAR dataset named Motion4D. This comprehensive dataset comprises $1,370$ 4D sequences totaling $124K$ frames, encompassing a wide variety of motion patterns. The diversity and scale of Motion4D significantly facilitate effective motion knowledge translation for SRMP, providing a robust foundation for both training and evaluating motion prediction models in a synthetic-to-real domain adaptation context. Based on Motion4D, we conduct a series of experiments to evaluate the effectiveness of SR-Motion. The results show that SR-Motion effectively enhances the motion regression branch, facilitating robust knowledge translation and further leading to superior performance on real scenes. 

Our contributions are three-fold:
\begin{itemize}
    \item We introduce the Synthetic-to-Real Motion Prediction (SRMP) task, the first formal study on transferring motion knowledge from synthetic to real 4D LiDAR data. We identify that motion jitters and object-level inconsistencies, caused by naive cell-wise regression under domain shift, are key barriers to robust knowledge transfer.
    \item We propose SR-Motion, a novel framework that leverages learned objectness priors to mitigate the sensitivity of motion regression under domain shift. The proposed objectness-aware prediction and objectness-aided enhancement modules offer a new paradigm for objectness-aware motion transfer.
    \item We present Motion4D, the first large-scale synthetic 4D LiDAR dataset tailored for SRMP, featuring diverse and realistic dynamic scenes and accurate motion annotations. Motion4D provides a benchmark for evaluating motion transfer methods and facilitates scalable training.
\end{itemize}

\section{Related Work}  \label{sec:related work}
\subsubsection{Motion Prediction on LiDAR Point Clouds}
Motion prediction~\cite{chai2020multipath,liang2020pnpnet,motionnet,wang2022sti,meng2023hybrid} aims to predict the future displacements of current point clouds based on past observations. To predict the motion efficiently, many recent works represent the 3D point cloud environment with bird's eye view (BEV) map and represent the motion as 2D displacement along the ground plane. Footbots~\cite{capellera2024footbots} propose a transformer-based architecture to predict motion trajectories for soccer players on BEV maps. PillarFlow~\cite{lee2020pillarflow} adopts PWC-Net~\cite{sun2018pwc} structure for 2D flow estimation to establish correspondences between two BEV maps. MotionNet~\cite{motionnet} and BE-STI~\cite{wang2022sti} learn to predict 2D motion in a fully-supervised manner by taking past BEV maps as input. SmartRefine~\cite{zhou2024smartrefine} proposes a scenario-adaptive refinement method to enhance motion prediction performance, but it still relies on expensive real-world HD maps and labeled trajectories.
Regarding the expensive cost of annotations, WeakMotionNet~\cite{li2023weakly} and SSMP~\cite{wang2024semi} propose to train the motion prediction networks in weakly- and semi-supervised manner, respectively. To be completely independent of annotations, PillarMotion~\cite{luo2021self} and SelfMotion~\cite{wang2024self} propose to learn from the data itself, training the model in a self-supervised manner.
Observing synthetic data can directly provide accurate motion labels, we propose to explore SRMP to acquire real-world motion knowledge from synthetic data.

\subsubsection{Synthetic-to-Real Translation}
Synthetic-to-real translation approaches~\cite{chen2020automated,gta-sf,lin2024synthetic,liu2021synthetic,liu2024textadapter,zeng2024hardness} aim to train a model on synthetic data and transfer the knowledge into real data, which is widely used to avoid the cost of laborious annotation of real data. To narrow the domain gap, some methods adopt representation learning to learn more transferable features across domains. For example, \cite{gao2020feature} aligns density map features to improve crowd counting in real scenes, \cite{gao2024audio} explores audio-visual representations for crowd anomaly detection, and \cite{li2024pre} learns general-purpose trajectory representations from synthetic data.

Recently, teacher-student frameworks~\cite{mt,wu2024instance,gou2024reciprocal} have become popular, where a teacher model generates pseudo-labels on real data to guide the synthetic-to-real learning of the student model. Most of the teacher-student-based works~\cite{yang2021st3d,yang2022st3d++,gta-sf,karim2023c,zhang2021prototypical,pan2024pseudo} focus on utilizing task-specific priors to improve the quality of pseudo labels and further alleviate the task-specific domain shifts. ST3D~\cite{yang2021st3d} introduces a quality-aware triplet memory bank to update pseudo labels to guide the knowledge transfer on the target domain. DCA-SRSFE~\cite{gta-sf} addresses the distortions and misalignments for 3D scene flow estimation.
However, SRMP is still unexplored, and there are no appropriate techniques for such a dense regression task in the 4D LiDAR sequence domain.
Thus, in this work, we concentrate on addressing domain gaps for SRMP to ensure a robust motion knowledge translation.

\subsubsection{Synthetic Dataset Generation} There are many works in multimedia application that generate large-scale synthetic datasets for improved and robust real-world performance. However, there is a lack of suitable synthetic datasets for SRMP.
A similar task with motion prediction is scene flow estimation, where synthetic datasets FT3D~\cite{ft3d} and GTA-SF~\cite{gta-sf} are widely used as benchmarks. FT3D focuses on general object movement. GTA-SF employs city engine GTA-V~\cite{gta-v} to simulate traffic and offer city scenes. For trajectory prediction, \cite{li2024pre} generates and refines synthetic trajectories on real-world HD maps to construct synthetic datasets.
While recent works~\cite{ramasinghe2020spectral,huang2022generation,zyrianov2025lidardm} have explored deep generative models to synthesize LiDAR data, these methods are prone to producing artifacts and risk model collapse through recursive training~\cite{shumailov2024ai}.
To circumvent these limitations, we adopt a physically-grounded simulation pipeline~\cite{lidar-aads,lidar-augmented,lidar-aug,lidar-blainder,lidar-sim,gta-sf,carla}. Instead of learning data distributions, our approach populates 3D environments with dynamic CAD assets and renders realistic 4D sequences. This is achieved through a high-fidelity simulator that employs ray-casting algorithms~\cite{cubvh,lidar-blainder} combined with z-buffer occlusion handling. The resulting dataset, Motion4D, features diverse and realistic traffic scenarios, providing a robust foundation for the development and evaluation of SRMP.

\begin{figure*}[!t]
\centering
\includegraphics[width=1.0\linewidth]{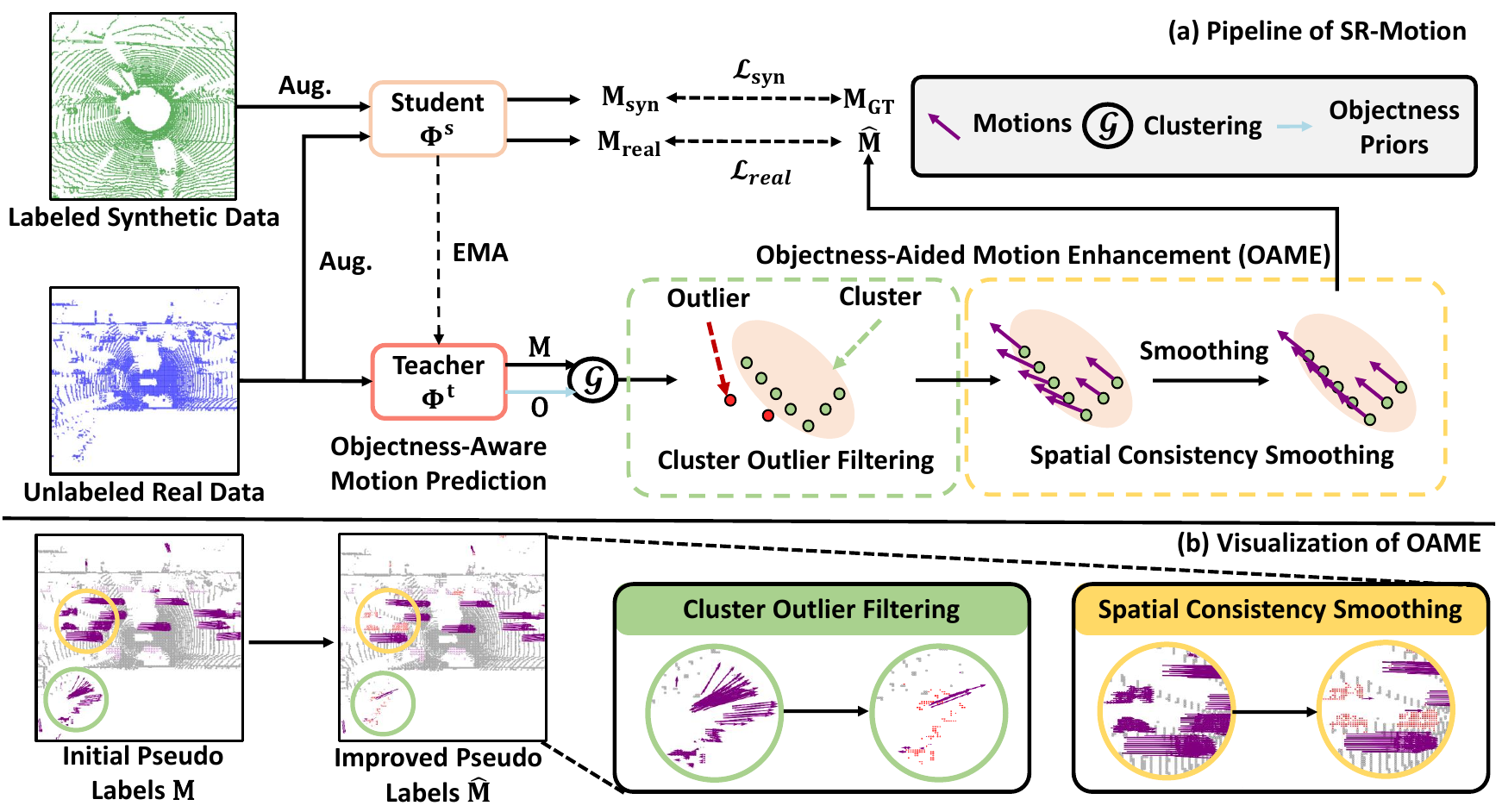}
\caption{\textbf{The pipeline of SR-Motion.} Here, we focus on the pivotal aspects of motion knowledge transfer for clarity. (a) The pipeline of our teacher-student-based SR-Motion. Both $\Phi^t$ and $\Phi^s$ are objectness-aware motion prediction networks (OAMNet), producing motions $M$ and objectness priors, i.e. centroid offsets $O$. The motions $M$ from $\Phi^t$ can be utilized as initial pseudo labels for real data. (b) OAME leverages objectness priors $O$ to generate improved pseudo labels $\hat{M}$. High-quality motion pseudo labels should exhibit consistency and smoothness within moving objects. OAME identifies and excludes noisy motion predictions depicted as orange cells, and smooths the distribution of motion predictions with objects, thereby refining the accuracy and reliability of the motion labels.}
\label{fig:pipeline}
\vspace{-0.4cm}
\end{figure*}

\section{Methodology} 

\subsection{Overview} \label{sec:overview}
In this work, we tackle the challenge of synthetic-to-real translation for motion prediction (SRMP), with the goal of leveraging synthetic data to perform accurate predictions on unlabeled real-world datasets. To achieve this objective, we address three pivotal questions: 1) How to effectively translate motion knowledge from synthetic to real environments? 2) How to perform robust motion prediction with synthetic-to-real domain gaps? 3) How to improve the quality and robustness of the translation process. 

As illustrated in Fig.~\ref{fig:pipeline}, our proposed solution, SR-Motion, is designed to address these challenges through a cohesive framework comprising three key components: motion knowledge translation framework in Sec~\ref{sec:se}, objectness-aware motion prediction in Sec~\ref{sec:model}, and objectness-aided motion enhancement in Sec~\ref{sec:pseudo-labeling}. These components work together to enable effective and robust knowledge translation from synthetic to real environments. In the subsequent sections, we will elaborate on each component of SR-Motion.

\subsection{Motion Knowledge Translation} \label{sec:se}
In the context of SRMP, a critical problem is the effective translation of motion knowledge from labeled synthetic data to unlabeled real-world data. As illustrated in Fig.~\ref{fig:pipeline}, SR-Motion adopts a teacher-student framework~\cite{mt} for foundational knowledge transfer. The key insight of this framework is to train a teacher model $\Phi^t$ using labeled synthetic data, which then generates high-quality pseudo labels $\hat{M}$ for real data to guide the learning of a student model $\Phi^s$. The training procedure is structured as follows:

\subsubsection{Teacher Model Initialization}
Initially, the teacher model $\Phi^t$ is trained on labeled synthetic data to capture motion knowledge effectively. Once adequately trained, the weights of $\Phi^t$ are copied to initialize the student model $\Phi^s$, ensuring that both models share identical architectures and start with the same learned parameters.

\subsubsection{Pseudo Label Generation}
Then, the teacher model $\Phi^t$ generates high-quality pseudo labels $\hat{M}$ for real-world data. These pseudo labels serve as a bridge between the synthetic and real domains, enabling $\Phi^s$ to learn from real-world scenarios without direct supervision. Besides, to mitigate the risk of noisy labels undermining robust knowledge translation, we enhance $\Phi^t$ with an objectness-aided motion enhancement module (detailed in Sec.~\ref{sec:pseudo-labeling}) to further improve the quality of pseudo label.

\subsubsection{Student Model Learning}
Obtained pseudo-labeled real-world data, $\Phi^s$ is able to jointly learn from losses $\mathcal{L}_{real}$ and $\mathcal{L}_{syn}$ on synthetic data and real data, respectively, via stochastic gradient descent. This learning strategy allows $\Phi^s$ to integrate motion knowledge from synthetic data while adapting to real-world conditions. The definitions of these two losses will be detailed in Sec.~\ref{sec:model}.

To boost the robustness of knowledge translation, data augmentation techniques are applied to the input sequences of $\Phi^s$. Specifically, we randomly apply the same rotation ($+90^{\circ}\ or -90^{\circ}$) and flip ($x$ or $y$ direction) to all frames in one sequence, with corresponding transformations applied to the (pseudo) labels to align with the modified inputs. This process ensures that motion prediction models produce consistent results despite various disturbances, thereby enhancing robustness. The implementation of augmentation is detailed in Sec.~\ref{sec:model}.

\subsubsection{Teacher Model Update}
Finally, to continuously refine the teacher model, after the $\tau_{th}$ iteration of student model learning, the teacher model $\Phi^t$ is updated according to the exponential moving average (EMA) of the student model $\Phi^s$:
\begin{equation}
\label{eq:ema}
\Phi^t_{\tau + 1} = \alpha \cdot \Phi^t_{\tau} + (1-\alpha) \cdot \Phi^s_{\tau + 1},
\end{equation}
where $\alpha$ is a smoothing factor to alleviate the noise in the single model $\Phi^s_\tau$. This update strategy effectively mitigates the impact of noisy or unstable gradients in the student model, leading to a more stable and robust teacher model over time. By gradually incorporating the learned representations of the student model $\Phi^s$, especially those adapted to the real-world domain through pseudo-labeling, the teacher $\Phi^t$ becomes increasingly better adapted to the target distribution. This results in higher-quality and more consistent pseudo-labels, which in turn provide more reliable supervision for the student in subsequent iterations.

Following each update, the knowledge translation process iterates back to the pseudo label generation step, allowing for continuous refinement and learning of both $\Phi^t$ and $\Phi^s$, progressively enhancing the performance of motion prediction. In the inference stage, the output of the motion branch of the teacher model is utilized for evaluation. 

\subsection{Objectness-Aware Motion Prediction} \label{sec:model}
\begin{figure}[!t]
\includegraphics[width=1.0\linewidth]{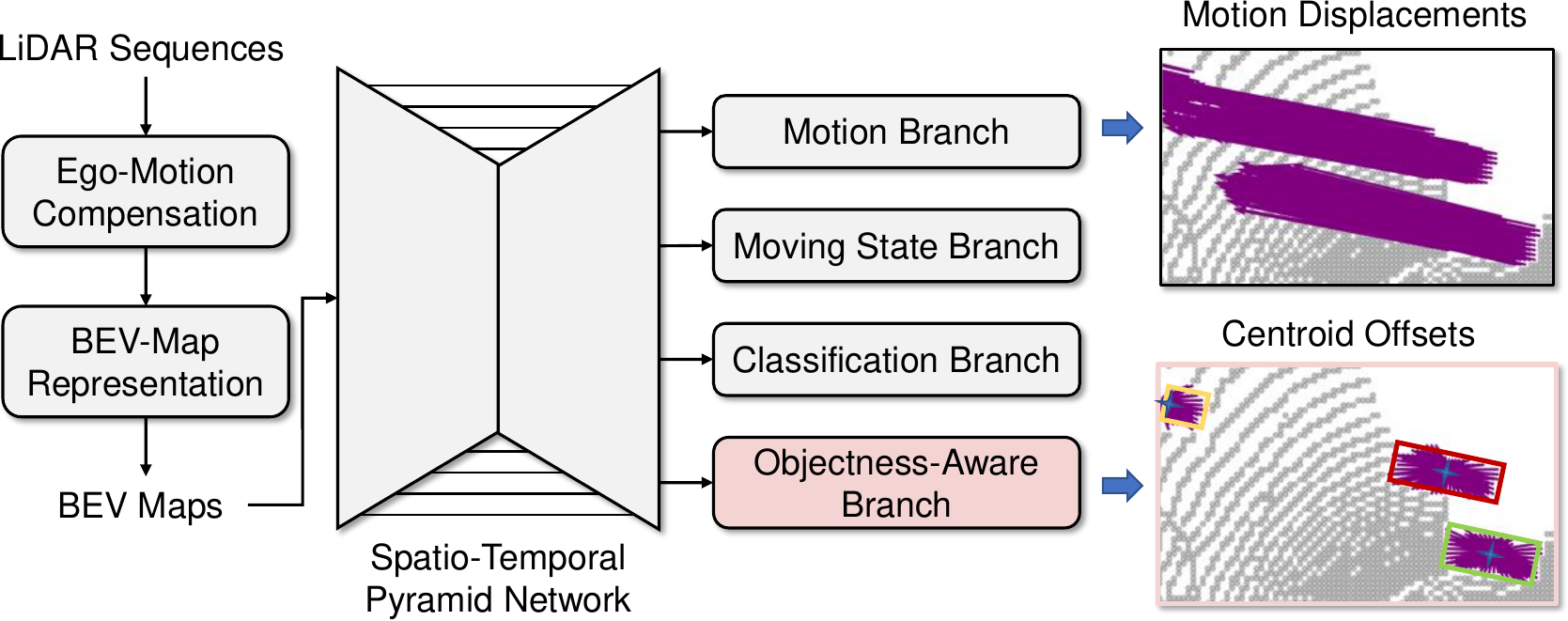}
\caption{\textbf{The architecture of OAMNet.} We introduce an objectness-aware branch to enhance the motion branch in a multi-task learning paradigm. Here, the purple arrows represent predictions of motion displacements and centroid offsets, respectively. We utilize the bounding boxes to highlight objects, with the stars denoting the centroids of the objects. The centroid offsets direct cells to the centroids of corresponding bounding boxes of objects. }
\label{fig:OAMNet}
\vspace{-0.6cm}
\end{figure}

While the above teacher-student framework has enabled basic motion knowledge translation, the simplistic motion regression method is sensitive to the domain shift, inevitably leading to noisy pseudo motion labels. To address this challenge, we introduce the Objectness-Aware Motion Prediction network (OAMNet), designed to improve the robustness of motion prediction across these domain gaps. As shown in Fig.~\ref{fig:OAMNet}, OAMNet builds upon the state-of-the-art MotionNet~\cite{motionnet} and enhances its capability to handle domain shifts by incorporating an objectness-aware branch into the architecture. We will provide a brief overview of the motion prediction pipeline from MotionNet, highlighting the feature extraction module and motion branch, while placing particular emphasis on the functionality of the newly introduced objectness-aware branch. We strongly recommend readers refer to MotionNet~\cite{motionnet} for details of feature extraction and motion branches.

\subsection{Preliminaries: From Point Clouds to BEV Representation}
This section details the nature of this data and the preprocessing steps used to create a structured input representation suitable for our deep learning architecture.

Our model processes 4D LiDAR data, which is a sequence of 3D LiDAR point clouds. A raw LiDAR point cloud is an unordered and sparse set of 3D points that captures the geometry of the surrounding environment. This format is not directly amenable to standard convolutional architectures. Therefore, we convert the sequence of raw point clouds into a dense, spatio-temporal tensor using a voxel-based, Bird's-Eye-View (BEV) representation, which is a common and effective approach~\cite{motionnet,li2024bevformer,yang2023bevformer,wang2024self} in autonomous driving perception. We follow the previous work to adopt a transformation process that involves the following key steps:
\begin{itemize}
    \item Temporal Alignment: Given an input sequence of $N$ LiDAR frames (the current frame and $N-1$ past frames), each of the frames is described by its local coordinate system. Thus, we first apply Ego-motion compensation to align all past point clouds to the coordinate system of the current frame. This crucial step corrects for the movement of the ego-vehicle and creates a coherent scene representation over time.
    \item Voxelization: The 3D space of the synchronized point clouds is quantized into a regular grid of voxels. We use a binary occupancy representation, where a voxel has a value of 1 if it contains one or more LiDAR points and 0 otherwise. The implementation details are presented in Sec.~\ref {sec:details}.
    \item BEV Pseudo-Image Formation: The 3D voxel grid is then projected onto a 2D BEV plane of size $H\times W$. To retain crucial vertical information, the height dimension is encoded into the feature channel. Specifically, for each grid cell in the BEV map, we form a binary vector of dimension $D$. Each element in this vector corresponds to a voxel at a specific height in that vertical column, indicating its occupancy status. This process transforms each LiDAR frame into a 2D pseudo-image where the channel dimension encodes height information. 
\end{itemize}
The final input to our network is a tensor with the shape $N\times H\times W\times D$, where $N$ is the number of temporal frames in the sequence, $H$ and $W$ are the height and width of the BEV grid, $D$ is the number of voxel layers along the vertical axis, now serving as the feature dimension. This structured tensor effectively captures the spatial and temporal dynamics of the driving scene in a form that can be further effectively processed.

\subsubsection{Motion Prediction}
Obtained the BEV representation, we adopt a Spatio-Temporal Pyramid Network (STPN)~\cite{motionnet} as the backbone architecture to process these input sequences. The SPTN hierarchically extracts features from a sequence of BEV maps by using efficient 2D convolutions for spatial information and 3D convolutions for temporal information. It then fuses these multi-scale features to combine both local and global spatio-temporal context for dense prediction tasks, resulting in extracted features $F\in\mathbb{R}^{N\times H\times W\times C}$ (where the output feature dimension $C=32$). The features are then forwarded to the following branches to perform various perception tasks.

As the pivotal branch for motion prediction, the motion branch, a regression head with two-layer 2D convolutions, predicts the motion vectors $M = \{M^t\}^T_{t=1}\in \mathbb{R}^{T\times H\times W \times 2}$ for each cell in the current BEV map. These motion vectors serve as offset indicators to forecast the future positions of cells in the subsequent $T$ frames, which can be mathematically expressed as:
\begin{equation}
\label{eq:motion}
X^t = X + M^t, \quad t=1,2,...,T,
\end{equation}
where $X\in \mathbb{R}^{H\times W \times 2}$ represents the coordinates of cells in the current BEV map, and $X^t\in \mathbb{R}^{H\times W \times 2}$ denotes their predicted coordinates in the $t^{th}$ future frame. The motion branch is supervised by a smooth L1 loss:
\begin{equation}
    \mathcal{L}_{m}=\sum_{t=1}^Tsmooth_{L1}(M^t, \overline{M}^t) \,,
\end{equation}
where $\overline{M} = \{\overline{M}^t\}^T_{t=1}\in \mathbb{R}^{T\times H\times W \times 2}$ is the ground truth motion vectors.

Additionally, following previous works~\cite{motionnet,wang2024semi}, two auxiliary branches are responsible for perceiving the categories and moving states of all cells, providing supplementary information for regularizing the motion predictions. Both of the branches are supervised by cross-entropy losses $\mathcal{L}_{p}$, which can be formualted as:
\begin{equation}
    \mathcal{L}_p=CELoss(A, \overline{A})+BCELoss(S,\overline{S})\,,
\end{equation}
where the first term is the Categorical Cross-Entropy loss for object classification. $A\in\mathbb{R}^{H\times W \times 5}$ contains the predicted probabilities for five categories (background, pedestrian, car, bike, and other), while  
$\overline{A}\in\mathbb{R}^{H\times W \times 5}$ is the corresponding one-hot ground-truth label. The second term is the Binary Cross-Entropy loss for moving state estimation. $S\in\mathbb{R}^{H\times W}$ is the predicted probability of an object being in motion, and $\overline{S}\in\mathbb{R}^{H\times W}$ is the ground-truth label indicating its state (static or moving).

\subsubsection{Objectness-Aware Branch}
Despite its effectiveness for single-domain motion prediction, naive cell-wise regression within the motion branch is sensitive to domain shifts between synthetic and real-world scenes, leading to noisy predictions (see Fig.~\ref{fig:pipeline}). Traditional approaches~\cite{motionnet} suggest constraining motions with rigid object labels, but this is non-trivial for SRMP due to the lack of labeled real data. To address this, OAMNet incorporates an objectness-aware branch with two advantages:
\begin{itemize}
    \item Implicit Integration: Via multi-task learning~\cite{caruana1997multitask,zhang2021survey}, the branch integrates learnt objectness knowledge with motion knowledge, enhancing prediction robustness.
    \item Explicit Enhancement: The branch predicts objectness priors, which can be utilized to constrain motions without requiring labels on real data (see Sec~\ref{sec:pseudo-labeling}).
\end{itemize}

\textbf{Detailed implementation.} Specifically, OAMNet adds the objectness-aware branch parallel to the motion branch. The objectness-aware branch is also implemented as a regression head with two-layer 2D convolutions. This design preserves the full encoder and context modules of MotionNet~\cite{motionnet} unchanged, ensuring backward compatibility while enriching motion prediction with objectness-aware cues.

The objectness-aware branch takes the spatio-temporal feature $F$ from the shared spatio-temporal pyramid network as input, which is identical to the motion branch. Inspired by detection and segmentation techniques~\cite{yin2021center,qi2019deep,jiang2020pointgroup,wu20223d,mei2023centerlps}, this branch predicts cell-wise relative offset vectors $O\in \mathbb{R}^{H\times W \times 2}$ that align cells with respective ground-truth centroids.
To this end, the branch consists of two convolutional layers:
\begin{itemize}
\item A $3\times3$ convolution with $32$ output channels, followed by BatchNorm and ReLU activation;
\item A $1\times1$ convolution with $2$ output channels (for $x$ and $y$ offsets), without non-linearity.
\end{itemize} 

During training on synthetic data, the ground-truth offset $\overline{O}$ for each cell at position $X=(x,y)$ is computed as $\overline{O}=X_{ctr}-X$, where $X_{ctr}\in\mathbb{R}^{H\times W \times 2}$ denotes the center coordinate of the bounding box enclosing the object which the cell belongs. Cells not associated with any object (i.e., background) are assigned zero offsets. The loss function combines L1 regression and cosine similarity to optimize length and direction jointly:
\begin{equation}
\label{eq:centroid}
\mathcal{L}_{o}=L1(O,\overline{O}) + Mean(\frac{O \cdot \overline{O}}{\Vert O \Vert \cdot \Vert \overline{O} \Vert})
\end{equation}
On real-world data, where ground-truth object labels are unavailable, we generate pseudo-offset labels using predictions from the teacher model $\Phi^t$(see Sec.~\ref{sec:se}), enabling self-supervised objectness learning.

By learning centroid offsets, the objectness-aware branch enriches OAMNet with objectness knowledge, i.e. enabling each cell to notice which instance it belongs to. This method encourages the motion branch to predict consistent motion for cells within the same instance, resulting in more reliable motion predictions.

\subsubsection{Losses and Augmentation}
For the synthetic domain, $\mathcal{L}_{syn}$ contains supervised losses from all
branches:
\begin{equation}
\mathcal{L}_{syn}=\mathcal{L}_{m}+\mathcal{L}_{p}+0.01\mathcal{L}_{o}.
\end{equation}
In contrast, for the real domain, $L_{real}$ only considers losses from motion and offset branches, mitigating impacts from the class imbalance, which can be formulated as:
\begin{equation}
\mathcal{L}_{real}=\mathcal{L}_{m}+0.01\mathcal{L}_{o}.
\end{equation}

In addition, to enhance the robustness of the model to geometric variations and domain gaps, we apply a consistent rotation and flip augmentation strategy on both the BEV input sequences and corresponding labels. The procedure is detailed in Algorithm~\ref{alg:augmentation}.

\begin{algorithm}[!t]
\caption{Rotation and Flip Augmentation}\label{alg:augmentation}  
\begin{algorithmic}[1]
\REQUIRE 
    Raw BEV sequence $ B = \{B^n\}_{n=1}^N $ with $ N $ past frames; \\
    Motion labels $ \overline{M} = \{{\overline{M}}^t\}_{t=1}^T$ for $ T $ future frames \\
    Offset labels $ \overline{O} = \{{\overline{O}}^t\}_{t=1}^T$ for $ T $ future frames
\ENSURE 
    Augmented BEV sequence $B$;\\
    Augmented motion labels $\overline{M}$;\\
    Augmented offset labels $\overline{O}$

\STATE /*\textit{ Sample consistent transformation} */
\STATE Randomly sample rotation angle:\\\qquad $ \theta \in \{0^\circ, 90^\circ, 180^\circ, 270^\circ\} $
\STATE Randomly decide flip:\\\qquad $ f_x \sim \text{Bernoulli}(0.5) $, $ f_y \sim \text{Bernoulli}(0.5) $

\STATE /* \textit{Compute transformation matrix} */
\STATE Rotation matrix: $ \mathbf{R} = \begin{bmatrix} \cos\theta & -\sin\theta \\ \sin\theta & \cos\theta \end{bmatrix} $
\STATE Flip scaling: $ s_x = (-1)^{f_x} $, $ s_y = (-1)^{f_y} $
\STATE Combined transform: $ \mathbf{T} = \begin{bmatrix} s_x & 0 \\ 0 & s_y \end{bmatrix} \mathbf{R} $

\STATE /* \textit{Apply consistent augmentation} */
\FOR{$ n = 1 $ to $ N $}
    \STATE $ B_T^n \leftarrow \text{Rotate}(B^n, \theta) $
    \IF{$ f_x = 1 $}
        \STATE Horizontal flip: $ B^n \leftarrow \text{Flip}(B^n, \text{axis}=1) $ 
    \ENDIF
    \IF{$ f_y = 1 $}
        \STATE Vertical flip: $ B^n \leftarrow \text{Flip}(B^n, \text{axis}=0) $ 
    \ENDIF
\ENDFOR
\STATE $ B = \{B^n\}_{n=1}^N $
\STATE /* \textit{Transform pseudo-labels} */
\FOR{$ t = 1 $ to $ T $}
    \STATE $ \overline{M}^t \leftarrow \overline{M}^t \cdot \mathbf{T} $, $ \overline{O}^t \leftarrow \overline{O}^t \cdot \mathbf{T} $
\ENDFOR
\STATE  $ \overline{M} = \{\overline{M}^t\}_{t=1}^T $, $ \overline{O} = \{\overline{O}^t\}_{t=1}^T $

\end{algorithmic}
\end{algorithm}

\begin{figure*}[!t]
\vspace{-0.2cm}
\centering
\subfloat[Objectness-Aided Motion Enhancement (OAME).]{\includegraphics[width=0.7\linewidth]{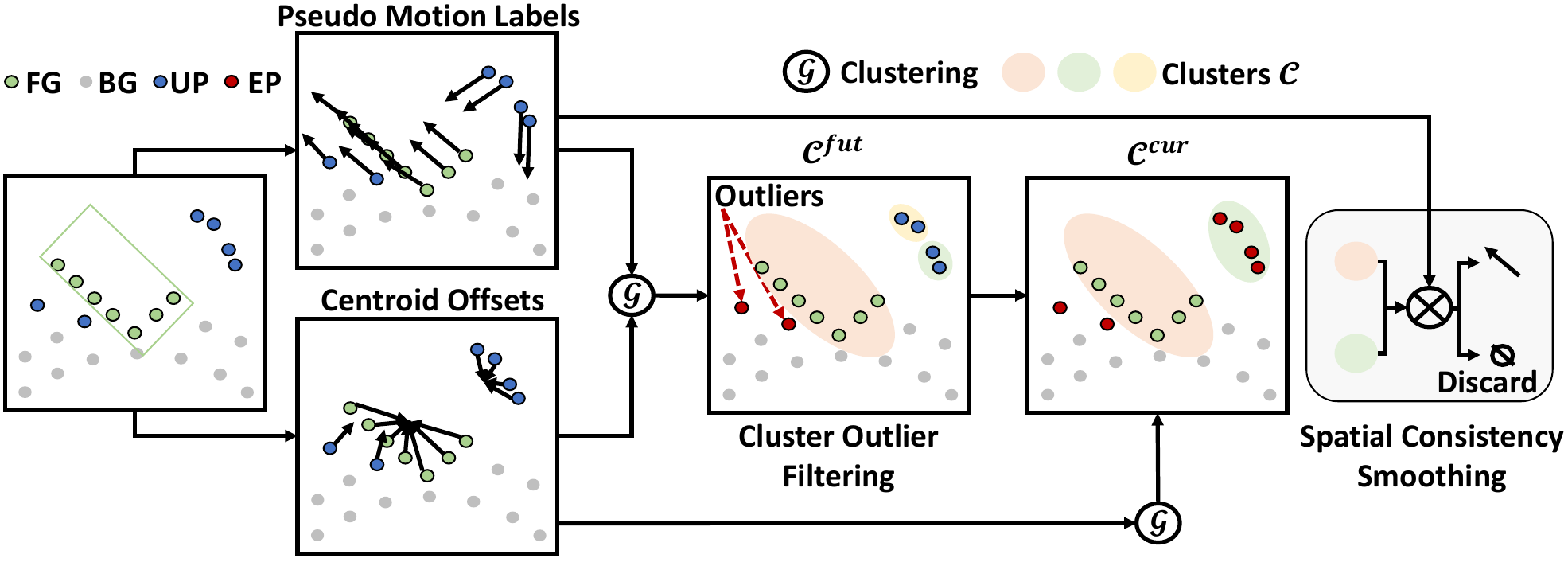}}
\hspace{1pt}
\rule[0cm]{1pt}{4.5cm}
\hspace{1pt}
\subfloat[Clustering.]{\includegraphics[width=0.18\linewidth]{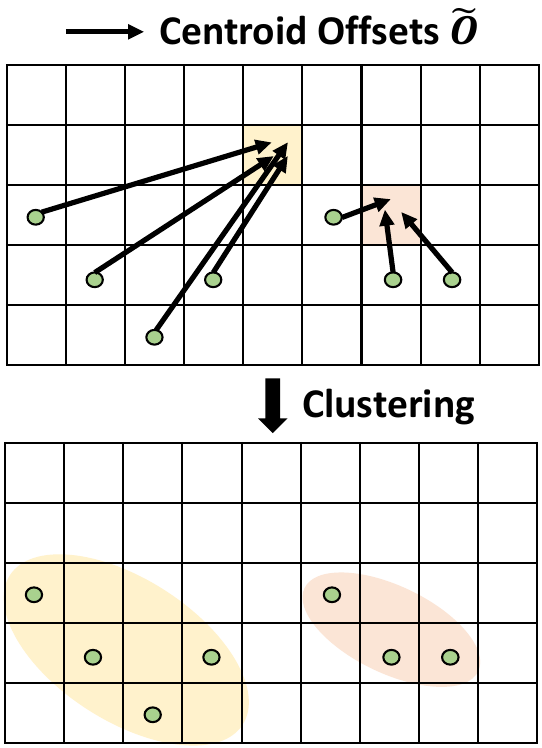}}
\caption{\textbf{The schematic of objectness-aided motion enhancement.} (a) We identify the unreliable pseudo labels (UP) and exclude these pseudo labels (EP). ``FG'' and ``BG'' denote foreground and background, respectively. (b) We present the centroid-aware clustering procedure. The cells sharing close shifted centroids are grouped into rigid clusters. }
\label{fig:refinement}
\vspace{-0.4cm}
\end{figure*}

\subsection{Objectness-Aided Motion Enhancement} \label{sec:pseudo-labeling}
Even with OAMNet implicitly enhanced robustness to domain gaps through multi-task learning, initial pseudo motion labels from the teacher model $\Phi^t$ still contain unavoidable noise, as illustrated in Fig.~\ref{fig:pipeline}. To mitigate the impact of this noise on knowledge transfer stability, we propose an Objectness-Aided Motion Enhancement (OAME) module that explicitly constrains these motion pseudo labels with objectness priors. The core principle underpinning OAME is the rigid assumption: \textit{cells within a rigid object should share consistent motions and common centroids}. As depicted in Fig.~\ref{fig:refinement}(a), the OAME module employs three strategies based on this assumption:
\begin{itemize}
    \item Centroid-aware clustering: Utilizing learned centroid offsets as cues to group cells into clusters that approximate rigid objects.
    \item Cluster outlier filtering (COF): Identifying and excluding noisy predictions that remain outliers in the clustering process.
    \item Spatial consistency smoothing (SCS): Ensuring consistent motion predictions for cells within the same cluster.
\end{itemize} 
Subsequently, we will detail the design of each component.

\subsubsection{Centroid-aware clustering} 
Given a set of centroid offset vectors $O$ and their corresponding cell coordinates $X$, we can calculate the centroid positions $X_{ctr} = X + O$. This operation effectively shifts each cell to its associated centroid location. If multiple cells have centroids that are very close or even identical, it is reasonable to assume these cells belong to the same rigid object. We then cluster these cells together based on their proximity in the transformed space. For example, as illustrated in Fig.~\ref{fig:refinement}(b), this process results in the scattered cells being grouped into two clusters, where cells within a cluster share identical centroid positions when mapped onto a quantized grid. We formulate the clustering process can be described as follows:
\begin{equation}
\mathcal{C} = \mathcal{G}(O, X, F),
\end{equation}
where  $\mathcal{C}$ denotes the set of all clusters, and $ \mathcal{G}$ is a function that takes the centroid offset vectors $O$, cell coordinates $X$, and a binary mask $F\in\mathbb{R}^{H\times W}$ indicating valid moving cells as inputs, and outputs the resulting clusters. For each cell $p$ within a cluster $\mathcal{C}_i$, we have:
\begin{equation}
[O_p + X_p] = X_{\mathcal{C}_i},
\end{equation}
where $[O_p + X_p]$ represents the integer coordinates obtained after applying the rounding function, and $X_{\mathcal{C}_i}$ is the shared centroid coordinate among all cells in the cluster $\mathcal{C}_i$.
This method aids in identifying objects that share similar spatial positions by grouping each instance into a cluster.

\subsubsection{Cluster outlier filtering (COF)}
Given the rigid assumption, cells within identical objects should maintain coherent clusters. Conversely, noisy cells with erratic motions or offsets cannot form valid clusters and will appear as outliers. Thus, we design COF method with two stages to suppress noisy pseudo labels with prediction jitters. 

In the first stage, we add the predicted cell-wise motions $M^T$ to cells to forecast their positions in the future $T^{th}$ frame, as described in Eq.~\ref{eq:motion}. The cells with motion jitters will be shifted to undesirable positions. Then, in the second stage, we add centroid offsets to determine the centroids of cells in future frames. Here, we can utilize the centroid-aware clustering technique to group moving cells into $\mathcal{C}^{fut}$. Thus, the overall procedure can be described by $\mathcal{C}^{fut} = \mathcal{G}(M^T + O, X, F)$. If any clusters in $\mathcal{C}^{fut}$ contain no more than $\mathcal{T}_{N}$ cells, they will be recognized as outliers, and therefore excluded from translation and labeled as invalid cells in $F$. 

\textbf{Discussion.} If we only activate the latter stage to cluster in the current frame, some noisy cells with wrong offsets, especially the ground cells near objects, may be incorporated into valid clusters. Thus, this one-stage filtering leads to inferior enhancement (see Table~\ref{table:ablation-COF}). In contrast, with the two-stage filter, noisy cells with either inaccurate motions or offsets will form invalid clusters, enabling us to filter out most of the unreliable jitters in pseudo labels.   

\subsubsection{Spatial consistency smoothing (SCS).}
Without explicit object constraints during motion regression, pseudo labels may exhibit inconsistent object-level motions, hindering stable knowledge transfer. Therefore, we design SCS to ensure the spatial consistency of motion across clusters.
Here, we first group the reserved valid cells from COF into clusters in current frames, which can be depicted as $\mathcal{C}^{cur} = \mathcal{G}(O, X, F)$. Since we have excluded the potential outliers via COF, these clusters are clean enough for consistency smoothing. For every cell $p$ belonging to $\mathcal{C}^{cur}_i$, we update its motion pseudo labels to the average motion $\mu M_{i}$ across $\mathcal{C}^{cur}_i$, which can be formulated as:
\begin{equation}
    \label{eq:cluster_mean}
    \forall p \in \mathcal{C}^{cur}_i\ M_p = \mu M_{i},\ \mu M_{i} = \frac{1}{N_i}\sum_{p\in \mathcal{C}^{cur}_i}{M_p},
\end{equation}
where $N_i$ denotes the number of cell in $\mathcal{C}^{cur}_i$.
Additionally, recognizing that clusters exhibiting high divergence in motion distribution are unreliable for smoothing and often contain significant noise, we introduce the Coefficient of Variation (CV) as a metric to evaluate the divergence within clusters. The CV of $\mathcal{C}^{s}_i$ can be defined as:
\begin{equation}
\begin{aligned}
    \label{eq:cluster_cv}
    CV_{i} = \frac{\sigma M_{i}}{\vert \mu M_{i} \vert},\ \sigma M_{i} = \sqrt{\frac{1}{N_i}\sum_{p\in \mathcal{C}^{cur}_i}{(M_p-\mu M_{i})^2}}.
\end{aligned}
\end{equation}
Here, $\sigma M_{i}$ represents the standard deviation of the motion values within cluster $\mathcal{C}^{s}_i$, while $\vert\mu M_{i}\vert$ denotes the absolute mean motion value of the same cluster. A higher $CV_{i}$ indicates greater variability and thus lower reliability of the cluster.

Clusters with a CV exceeding a predefined threshold $\mathcal{T}_{CV}$ in either the horizontal or vertical direction along the ground plane are considered unreliable and are discarded. This filtering process ensures more stable self-training by eliminating clusters that could otherwise introduce noise and instability into the model. By doing so, we enhance the robustness and accuracy of our motion prediction framework, particularly in scenarios characterized by significant domain shifts.

\def\thickhline{\noalign{\hrule height.8pt}}

\section{Motion4D Dataset} \label{sec:Motion4D}
We create a large-scale synthetic 4D dataset Motion4D with diverse labeled motions, focusing on addressing the lack of synthetic datasets in SRMP. In this section, we present the motion synthesis pipeline of Motion4D and its key features.

\subsection{Motion Synthesis Pipeline}\label{sec:pipeline}
Motion4D is built upon Waymo Open Dataset~\cite{waymo-motion, waymo-perception} and Blender~\cite{blender}-based LiDAR tracing engine BLAINDER~\cite{lidar-blainder}. Specifically, it synthesizes a sequence of realistic driving scenes by placing diverse moving objects onto a static background, which is a single static real-world LiDAR point cloud. 

We first conduct scene library and motion library to provide backgrounds and objects, respectively.
\textbf{Scene library} The real-world LiDAR point clouds with few foreground objects are suitable to serve as the background for motion placement and can be easily collected. Here, for efficiency and versatility, we select $685$ discrete static LiDAR frames collected from top center LiDARs in Waymo Perception Dataset~\cite{waymo-perception}. Note that the selected frames only serve as backgrounds without any annotations. These frames are expected to contain broad road areas to provide vast space and freedom for the generation of motion. Besides, we implement Patchwork++~\cite{patchwork++} for unsupervised road segmentation to judge valid areas of motion placement. 
\textbf{Motion library} Then, we establish the motion library by combining CAD assets with diverse trajectories from the training split of Waymo Motion Dataset~\cite{waymo-motion}. The CAD assets are adopted from banks in CARLA~\cite{carla} and Render People~\cite{3d-people}. For realism, we apply category-specific speed restrictions and rule-based invalid motion suppression.
Please refer to supplementary materials for details.

Now, with the above libraries, we can generate synthetic data via LiDAR simulation in BLAINDER. As the previous motion prediction approach~\cite{motionnet} proves that sweep synchronization affects the model performance greatly, we eliminate the ego-vehicle motions and freeze the backgrounds so that we can obtain naturally synchronized sequences. In the simulation, we first obtain rays in LiDAR-axes from static LiDAR backgrounds. Then, we freely place driving objects onto the valid area in the backgrounds. To improve realism, we also implement collision detection to regenerate invalid motions. 
Finally, to collect LiDAR sequences, we activate the placed driving objects by moving objects along the respective trajectories and iteratively re-scan the rays based on z-buffer occlusion at a predefined frequency (10 Hz). The multi-iteration ray tracing in BLAINDER ensures realistic simulation. In order to annotate motions, following the rigidity assumption~\cite{motionnet,li2023weakly,gta-sf}, we calculate motion displacements depending on object boxes and trajectories. Further details of LiDAR scanning are included in supplementary materials.

\subsection{Dataset Features} \label{sec:features}
\begin{figure}[!t]
\centering
\includegraphics[width=1\linewidth]{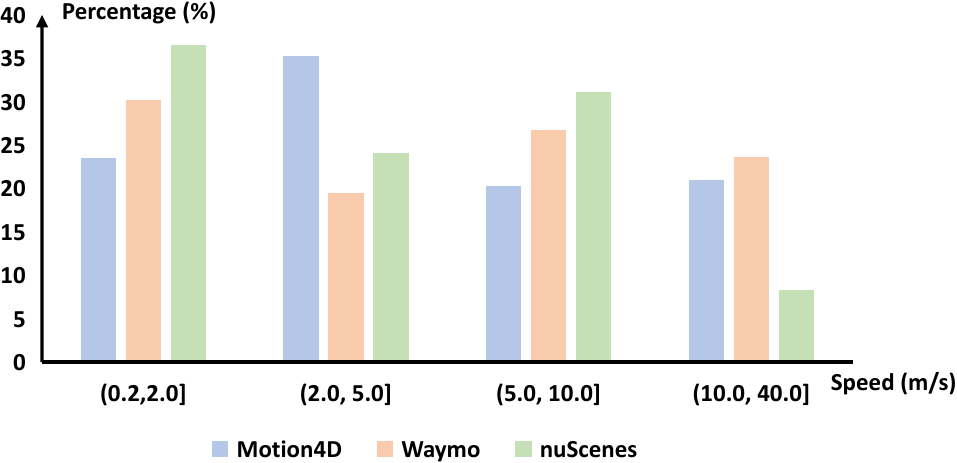}
\caption{\textbf{Distribution of motion speeds in real-world datasets and our synthetic Motion4D datasets.} Motions are categorized into four speed intervals, with percentages indicating the proportion of motions within each speed range relative to the total in the respective dataset.}
\label{fig:dist}
\vspace{-0.4cm}
\end{figure}

With the above pipeline, Motion4D is constructed as a large-scale 4D synthetic dataset specifically designed for motion prediction, which contains $1370$ sequences with $124K$ frames of LiDAR point clouds. Our careful design makes the dataset both realistic and cost-effective.
Each sequence in Motion4D provides diverse motions with precise labels based on just a single static LiDAR point cloud. Table~\ref{table:dataset-comparison} shows the comparison between real-world datasets and our synthetic datasets, demonstrating the diversity of motions and objects in Motion4D. As depicted in Fig.~\ref{fig:dist}, Motion4D exhibits a rich variety and balanced distribution of motions. Thus, serving as the synthetic source data, Motion4D lays a strong foundation for synthetic-to-real translation for motion prediction.

\begin{figure}
\centering
\includegraphics[width=0.9\linewidth]{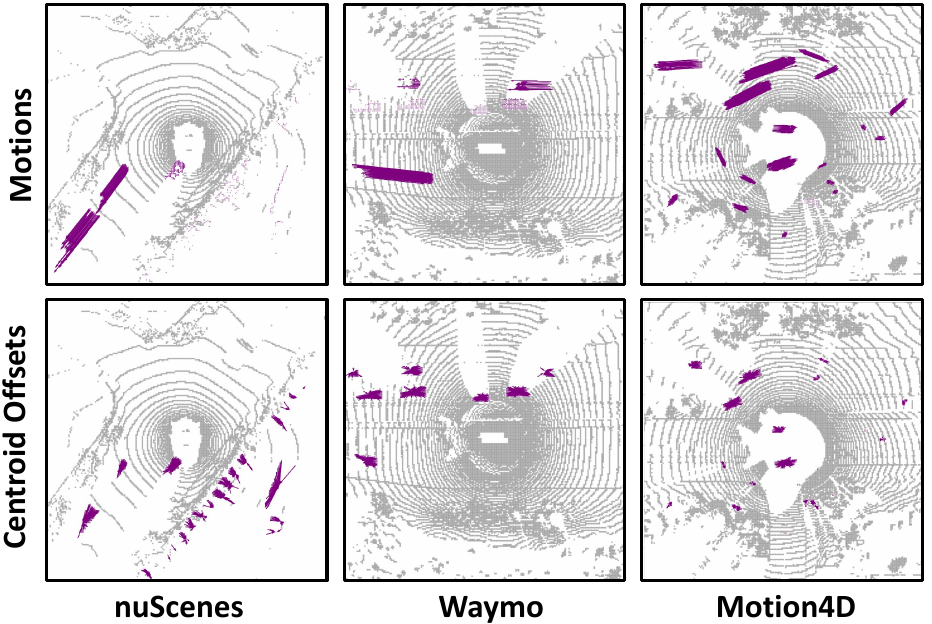}
\caption{\textbf{Visual comparisons of real-world datasets nuScenes~\cite{caesar2020nuscenes} and Waymo~\cite{waymo-perception}; and Motion4D.} Motion4D provides rich motion and object knowledge from diverse moving objects.}
\label{fig:datasets}
\vspace{-0.4cm}
\end{figure}
\begin{table}[!t]
    \centering
    \caption{\textbf{Comparison between related datasets and Motion4D.} We consider whether datasets provide synthetic, traffic scenes, and 4D sequences. Motion labels denote the average number of cell-wise labels per training sample.}
    \label{table:dataset-comparison}
    \resizebox{\linewidth}{!}{%
    \begin{tabular}{@{}c|ccc|cc@{}}
    \toprule
    \multirow{2}{*}{Datasets} & \multicolumn{3}{c|}{Features} & Training & Motion Labels \\
                             & Synthetic & Traffic & Sequences & Sequences & (1/sample) \\
    \midrule
    Waymo~\cite{waymo-perception} & \ding{55} & \checkmark & \checkmark & 798 & 1243.3 \\
    nuScenes~\cite{caesar2020nuscenes} & \ding{55} & \checkmark & \checkmark & 500 & 432.2 \\
    \midrule
    FT3D~\cite{ft3d} & \checkmark & \ding{55} & \ding{55} & - & - \\
    GTA-SF~\cite{gta-sf} & \checkmark & \checkmark & \ding{55} & - & - \\
    \textbf{Motion4D} & \checkmark & \checkmark & \checkmark & 1370 & 1779.6 \\
    \bottomrule
    \end{tabular}%
    }%
    \vspace{-0.4cm}
\end{table}
\begin{figure}[!t]
\centering
\includegraphics[width=0.8\linewidth]{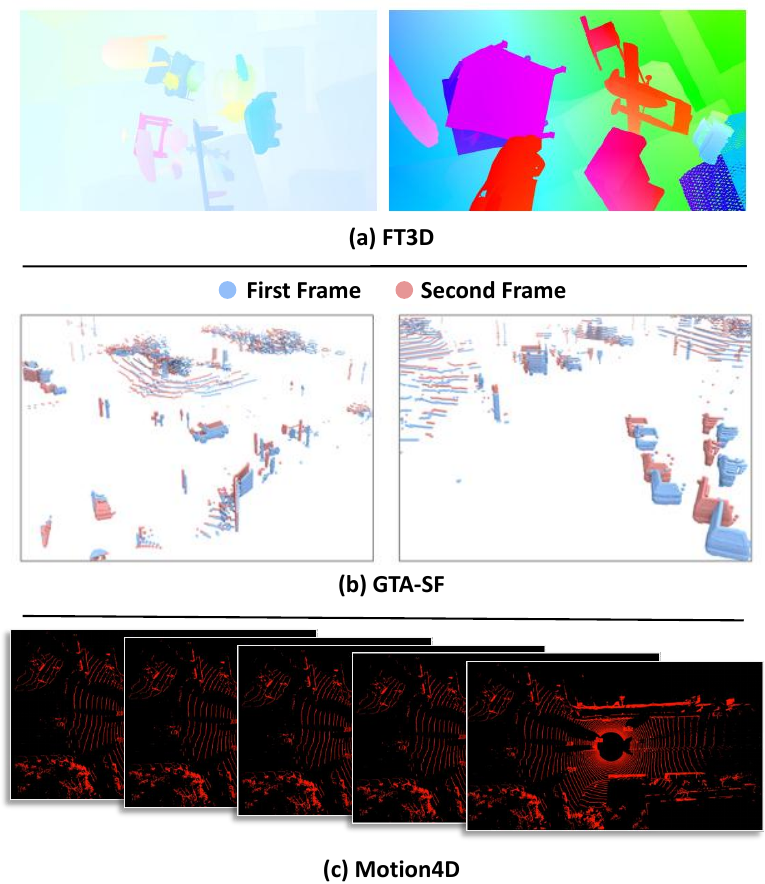}
\caption{\textbf{Visual comparisons of other synthetic datasets and Motion4D.} Both GTA-SF~\cite{gta-sf} and FT3D~\cite{ft3d} are datasets designed for scene flow estimation. For comparison, we show the first and second frames of two sequences from GTA-SF, two optical flow maps from FT3D, and a clip of an 4D sequence from Motion4D.}
\label{fig:vis-compare-syn-data}
\vspace{-0.5cm}
\end{figure}
A similar task with motion prediction is scene flow estimation, where synthetic datasets FT3D~\cite{ft3d} and GTA-SF~\cite{gta-sf} are widely used as benchmarks. We present a visual comparison of FT3D, GTA-SF, and Motion4D as shown in Fig.~\ref{fig:vis-compare-syn-data}. FT3D focuses on general object movement, which is unsuitable for autonomous driving scenarios. GTA-SF employs city engine GTA-V~\cite{gta-v} to simulate traffic and offer city scenes. 
However, it is worth noting that both datasets currently offer 3D point cloud pairs designed for scene flow estimation. While valuable for such estimation task, it may be considered insufficient for addressing the scope of motion prediction, which involves forecasting future motion displacement based on a comprehensive 4D sequence input. Besides, as for LiDAR simulation, FT3D and GTA-SF transform z-buffer to point clouds and need special designs to rectify the points. Motion4D considers refraction and reflection of lights and implements multi-iteration ray tracing to calculate realistic light paths.

\section{Experiments}
\subsection{Implementation Details}\label{sec:details}
\subsubsection{Real-world datasets}
We evaluate SR-Motion on two large-scale real-world autonomous driving datasets, Waymo~\cite{waymo-perception} and nuScenes~\cite{caesar2020nuscenes}. For Waymo, we follow WeakMotionNet~\cite{li2023weakly} to use $798$ scenes for training and $202$ scenes for validation. 
Note that we collect data from six sensors for a fair comparison. 
For nuScenes, we follow MotionNet~\cite{motionnet} to use $500$ scenes for training, $100$ scenes for validation, and $250$ scenes for testing. The motion ground truths are obtained from detection and tracking annotations~\cite{motionnet,li2023weakly}. 

\subsubsection{Training details} Following the same data preprocessing in previous works~\cite{motionnet,li2023weakly}, we crop the point clouds in the range of $[-32, 32]\times [-32, 32]\times [-1, 4]$ and $[-32, 32]\times [-32, 32]\times [-3, 2]$ meters for Waymo and nuScenes, respectively. The voxel size is set to $(0.25,0.25,0.4)m$. The past four sweeps are synchronized and combined with the current sweep as input. The network is trained on a single RTX 4090 GPU. The batch size is set to $16$ with $8$ synthetic data and $8$ real data. By operating directly in the Bird's Eye View (BEV) space, our architecture achieves high computational efficiency. During inference, it runs at an average speed of 20 ms per frame on the same RTX 4090, enabling real-time performance. Training typically takes approximately 5 hours.

\begin{table*}[!t]
\begin{center}
\caption{\textbf{Quantitative results on Waymo and nuScenes datasets under synthetic-to-real settings.} The superscripts $\dagger$ mean that the models are trained with only labeled synthetic data, and $\S$ mean that the models are trained in the mean teacher~\cite{mt} paradigm. It is advisable to evaluate models across three groups. Note that ``-'' denotes unreported results in the original work.}
\label{table:performance}
\resizebox{0.9\linewidth}{!}{\small
\begin{tabular}{@{}c|l|cc|cccccc@{}}
\toprule
\multirow{2}*{Real Data}&\multirow{2}*{Approach}&\multicolumn{2}{c|}{Supervision}&\multicolumn{2}{c}{Static}&\multicolumn{2}{c}{Speed $\leq$ 5m/s (slow)}&\multicolumn{2}{c}{Speed $\geq$ 5m/s (fast)}\\
~&~&Real&Synthetic&Mean$\downarrow$ &Median$\downarrow$ &Mean$\downarrow$ &Median$\downarrow$ &Mean$\downarrow$ &Median$\downarrow$\\
\midrule
\multirow{10}*{Waymo~\cite{waymo-perception}}&Oracle (MotionNet~\cite{motionnet})&\checkmark&-&0.025&0.0&0.261&0.085&0.994&0.539\\
~&WeakMotionNet~\cite{li2023weakly}&\checkmark&-&0.033&-&0.346&-&1.566&-\\
\cmidrule(r){2-10}
~&SelfMotion~\cite{wang2024self}&-&-&0.052&-&0.444&-&2.369&-\\
~&MotionNet$^\dagger$~\cite{motionnet}&-&\checkmark&0.081&0.0&0.486&0.277&2.398&0.929\\
~&MotionNet$^\S$~\cite{motionnet}&-&\checkmark&0.122&0.0&0.379&0.111&2.702&1.586\\
~&GRL~\cite{ganin2015unsupervised}&-&\checkmark&0.105&0.0&0.396&0.107&2.135&1.304\\
~&GU~\cite{zhang2023glenet}&-&\checkmark&0.068&0.0&\textbf{0.372}&0.105&2.388&0.976\\
\cmidrule(r){2-10}
~&Ours (OAMNet$^\dagger$)&-&\checkmark&\textbf{0.048}&0.0&0.502&0.314&2.383&0.935\\
~&Ours (OAMNet$^\S$)&-&\checkmark&0.049&0.0&0.418&\textbf{0.099}&2.010&1.192\\
~&Ours (OAMNet$^\S$+OAME)&-&\checkmark&\textbf{0.048}&0.0&0.385&\textbf{0.099}&\textbf{1.545}&\textbf{0.791}\\
\hline \hline
\multirow{11}*{nuScenes~\cite{caesar2020nuscenes}}&Oracle (MotionNet~\cite{motionnet})&\checkmark&-&0.026&0.0&0.257&0.096&1.074&0.733\\
~&WeakMotionNet~\cite{li2023weakly}&\checkmark&-&0.058&0.0&0.434&0.131&1.782&1.089\\
~&SSMP~\cite{wang2024semi}&\checkmark&-&0.015&0.0&0.350&0.102&1.941&1.217\\
\cmidrule(r){2-10}
~&PillarMotion~\cite{luo2021self}&-&-&0.162&0.001&0.697&0.176&3.550&2.084\\
~&MotionNet$^\dagger$~\cite{motionnet}&-&\checkmark&0.222&0.0&0.724&0.233&2.582&1.399\\
~&MotionNet$^\S$~\cite{motionnet}&-&\checkmark&0.450&0.0&0.700&0.213&2.248&1.251\\
~&GRL~\cite{ganin2015unsupervised}&-&\checkmark&0.284&0.0&0.571&0.145&2.519&1.423\\
~&GU~\cite{zhang2023glenet}&-&\checkmark&0.291&0.0&0.564&0.203&2.748&1.751\\
\cmidrule(r){2-10}
~&Ours (OAMNet$^\dagger$)&-&\checkmark&0.146&0.0&0.638&0.213&3.042&1.542\\
~&Ours (OAMNet$^\S$)&-&\checkmark&0.238&0.0&0.547&0.135&\textbf{2.241}&1.213\\
~&Ours (OAMNet$^\S$+OAME)&-&\checkmark&\textbf{0.062}&0.0&\textbf{0.421}&\textbf{0.112}&2.486&\textbf{1.167}\\
\bottomrule
\end{tabular}}
\end{center}
\vspace{-0.4cm}
\end{table*}

\subsubsection{Evaluation metrics} Following previous works~\cite{motionnet,li2023weakly}, each grid cell is categorized into one of three groups based on ground truth speeds: static, slow ($\leq 5m/s$), and fast ($\geq 5m/s$). This classification facilitates a more nuanced evaluation of model performance across different motion dynamics. To assess the accuracy of our predictions, we compute both the mean and median errors. These errors are quantified using the L2 distance between the predicted displacements and the corresponding ground truth displacements over a future time horizon of 1 second. Median values are supplementary, and mean metrics are used for primary comparison.

\subsubsection{Comparison methods} We present the comparison with the following methods. (a) \textit{Synthetic-only} models (labeled with superscripts $\dagger$) trained with only synthetic data; (b) \textit{Adaptation baseline} models (labeled with superscripts $\S$) trained in mean teacher paradigm to realize basic knowledge transfer; (c) \textit{General domain adaptation techniques}. Specifically, the gradient reversal layer (GRL)~\cite{ganin2015unsupervised} reverses the backpropagation of domain classification to render features invariant to domain shifts, and Gaussian uncertainty estimation (GU)~\cite{zhang2023glenet} predicts motions as Gaussian distributions and utilize the variances as uncertainty to exclude noisy pseudo labels; (d) Fully-~\cite{motionnet}, weakly-~\cite{li2023weakly}, semi-~\cite{wang2024semi} and self-supervised~\cite{luo2021self} motion prediction approaches. MotionNet~\cite{motionnet} trained with labeled real data is employed as \textit{Oracle} models for clear comparison.

\subsection{Main Results} \label{sec:main-results}
In Table~\ref{table:performance}, we evaluate and compare our SR-Motion on different real datasets: Waymo and nuScenes.

Firstly, we evaluated \textit{synthetic-only} approaches, specifically ``MotionNet$^\dagger$'' and ``OAMNet$^\dagger$''. Both methods demonstrated respectable performance on real-world datasets, confirming that our Motion4D dataset effectively simulates realistic traffic scenarios and provides a solid foundation for SRMP. Subsequently, we examined the baseline model ``MotionNet$^\S$'', which fundamentally transfers knowledge and slightly improves the prediction accuracy for slow cells. However, the presence of noisy pseudo labels inevitably degrades the performance for static and fast-moving cells. 

To mitigate this issue, we try to directly incorporate general adaptation techniques such as Gradient Reversal Layer (GRL)~\cite{ganin2015unsupervised} and Gaussian Uncertainty (GU)~\cite{zhang2023glenet} methods to enhance the baseline models. Although these adaptations improve performance, they still yield sub-optimal results. In contrast, SR-Motion introduces innovative components, OAMNet and OAME modules, which effectively bridge domain gaps for SRMP. Specifically, compared to the baseline model, SR-Motion reduces the mean error for the fast group from $2.702$ to $1.545$ and closes the performance gap by $76.3\%$ for the static group on the Waymo dataset.

Furthermore, we compared SR-Motion with \textit{Oracle}, weakly-, semi-, and self-supervised motion prediction approaches. Our method achieves comparable performance to fully-supervised \textit{Oracle} models. When compared with weakly- and semi-supervised approaches like WeakMotionNet and SSMP, our approach matches their performance without requiring any target data labels. Moreover, SR-Motion significantly outperforms previous self-supervised methods, PillarMotion and SelfMotion, narrowing the performance gap up to $86.4\%$ at the static speed level.

\noindent \textbf{Real-to-real Domain Adaptation}
In Table~\ref{table:domain-adaptation}, we present the results of real-to-real domain adaptation for motion prediction. On both Waymo and nuScenes datasets, our approach surpasses the baseline in real-to-real domain adaptation, affirming the effectiveness and robustness of SR-Motion. 

Moreover, models trained on our synthetic dataset, Motion4D, achieve performance comparable to their real-to-real counterparts across various adaptation settings, highlighting the strong generalization capability of Motion4D. Notably, in the ${Motion4D}\rightarrow{Waymo}$ setting, our approach significantly outperforms the ${nuScenes}\rightarrow{Waymo}$ setting, achieving a mean error of $1.545$ versus $2.327$ for the fast-moving agent group. The marked performance gain underscores the high realism and robust domain adaptability of the Motion4D dataset, facilitating accurate and generalizable motion prediction.

Furthermore, we trained MotionNet on a mixed dataset comprising both the original Waymo data and our synthetic Motion4D data. As shown in Table.~\ref{table:aug}, this model outperforms MotionNet trained exclusively on the Waymo dataset by a significant margin of $9.4\%$. This performance gain demonstrates that Motion4D serves as a valuable complementary data source that enriches the training distribution with diverse and realistic scenarios.

\begin{table}[!t]
\begin{center}
\caption{\textbf{Comparisons between different domain adaptation settings.} ${nuScenes}\rightarrow{Waymo}$ and ${Waymo}\rightarrow{nuScenes}$ are real-to-real settings. In contrast, experiments with Motion4D as the source domain are synthetic-to-real settings. $\dagger$ means that models are trained with only source data.}
\label{table:domain-adaptation}
\resizebox{\linewidth}{!}{
\begin{tabular}{@{}llc|ccc@{}}
\toprule
Source&Target&Approach&Static&Slow&Fast\\
\midrule

\multirow{2}*{nuScenes}&\multirow{4}*{Waymo}&MotionNet$^\dagger$&\textbf{0.021}&0.442&3.808\\
~&~&Ours&0.043&\textbf{0.377}&2.327\\
 \multirow{2}{*}{Motion4D} & ~ &  MotionNet$^\dagger$ & \cellcolor{gray!30} 0.081 & \cellcolor{gray!30} 0.486 & \cellcolor{gray!30} 2.398 \\
~ & ~ &  Ours & \cellcolor{gray!30} 0.048 & \cellcolor{gray!30}0.385 & \cellcolor{gray!30} \textbf{1.545} \\
\midrule
\multirow{2}*{Waymo}&\multirow{4}*{nuScenes}&MotionNet$^\dagger$&\textbf{0.040}&0.381&2.371\\
~&~&Ours&0.041&\textbf{0.319}&\textbf{1.611}\\
\multirow{2}*{Motion4D}&~&MotionNet$^\dagger$&\cellcolor{gray!30}0.222&\cellcolor{gray!30}0.724&\cellcolor{gray!30}2.582\\
~&~&Ours&\cellcolor{gray!30}0.062&\cellcolor{gray!30}0.421&\cellcolor{gray!30}2.486\\
\bottomrule
\end{tabular}}
\end{center}
\vspace{-0.2cm}
\end{table}

\begin{table}[!t]
\begin{center}
\caption{\textbf{Effectiveness of Motion4D as a complementary training set.} Performance is reported on the Waymo validation set, where lower is better ($\downarrow$).}
\label{table:aug}
\begin{tabular*}{0.9\linewidth}{@{}@{\extracolsep{\fill}}cc|ccc@{}}
\toprule
\multicolumn{2}{c|}{Training Source}&\multirow{2}*{Static$\downarrow$}&\multirow{2}*{Slow$\downarrow$}&\multirow{2}*{Fast$\downarrow$}\\
Motion4D&Waymo&~&~&~\\
\midrule
\checkmark& & 0.081 & 0.486 & 2.398 \\
&\checkmark &0.025&0.261 &0.994 \\
\checkmark&\checkmark&\textbf{0.025}&\textbf{0.247} &\textbf{0.901} \\
\bottomrule
\end{tabular*}
\end{center}
\vspace{-0.2cm}
\end{table}

\begin{figure*}[!t]
\centering
\includegraphics[width=0.95\linewidth]{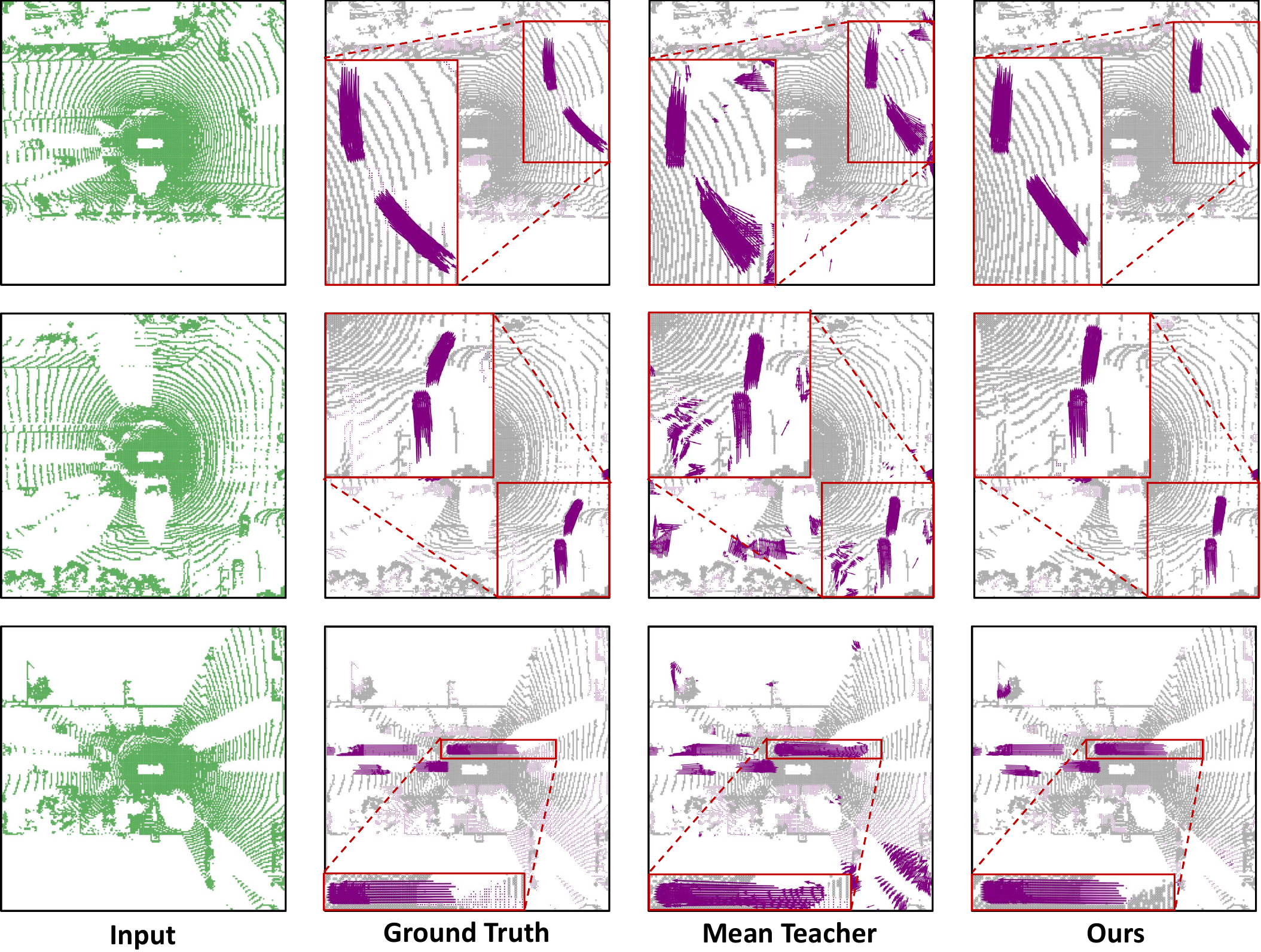}
\caption{\textbf{Visualization of motion prediction results on Waymo validation set.} ``Mean teacher'' is the adaptation baseline combining MotionNet with the basic teacher-student framework, Mean Teacher~\cite{mt}. Our approach shows more stable translation results.}
\label{fig:supp-vis}
\vspace{-0.0cm}
\end{figure*}

\subsection{Ablation Study}
We conduct ablation studies to evaluate the effectiveness of different components in SR-Motion. All the experiments are conducted on Motion4D$\rightarrow$Waymo setting and evaluated on Waymo val sets. Mean errors are reported.

\subsubsection{Evaluation of Different Components} 
SR-Motion comprises several key components: motion knowledge translation (MT), objectness-aware motion prediction (OA); objectness-aided motion enhancement with cluster outlier filtering (COF) and spatial consistency smoothing (SCS). As shown in Table~\ref{table:ablation}, we evaluate their contributions to SR-Motion.  

\begin{table}[!t]

\begin{center}
\caption{\textbf{Ablation on components in SR-Motion.}} 
\label{table:ablation}
\resizebox{0.9\linewidth}{!}{\small
\begin{tabular*}{1\linewidth}{@{}@{\extracolsep{\fill}}c|cccc|ccc@{}}
\toprule
{Group}&OA&MT&COF&SCS&Static&Slow&Fast\\
\midrule
Oracle & & & & &0.025&0.261 &0.994\\
\midrule
\romannumeral1 & & & & &0.081&0.486 &2.398 \\
\romannumeral2 &\checkmark & & & &0.048&0.502 &2.383 \\
\midrule
\romannumeral3  & &\checkmark & & &0.122&0.379 &2.701 \\
\romannumeral4  &\checkmark &\checkmark & & &0.049&0.418 &2.010 \\
\romannumeral5  &\checkmark &\checkmark &\checkmark & &0.035&0.410 &1.936 \\
\romannumeral6  &\checkmark &\checkmark & &\checkmark &0.068&0.387 &2.174 \\
\romannumeral7  &\checkmark &\checkmark &\checkmark &\checkmark & 0.048 & 0.385 & 1.545  \\
\bottomrule
\end{tabular*}}
\end{center}
\vspace{-0.6cm}
\end{table}
\begin{figure*}[!t]
\centering
\includegraphics[width=0.95\linewidth]{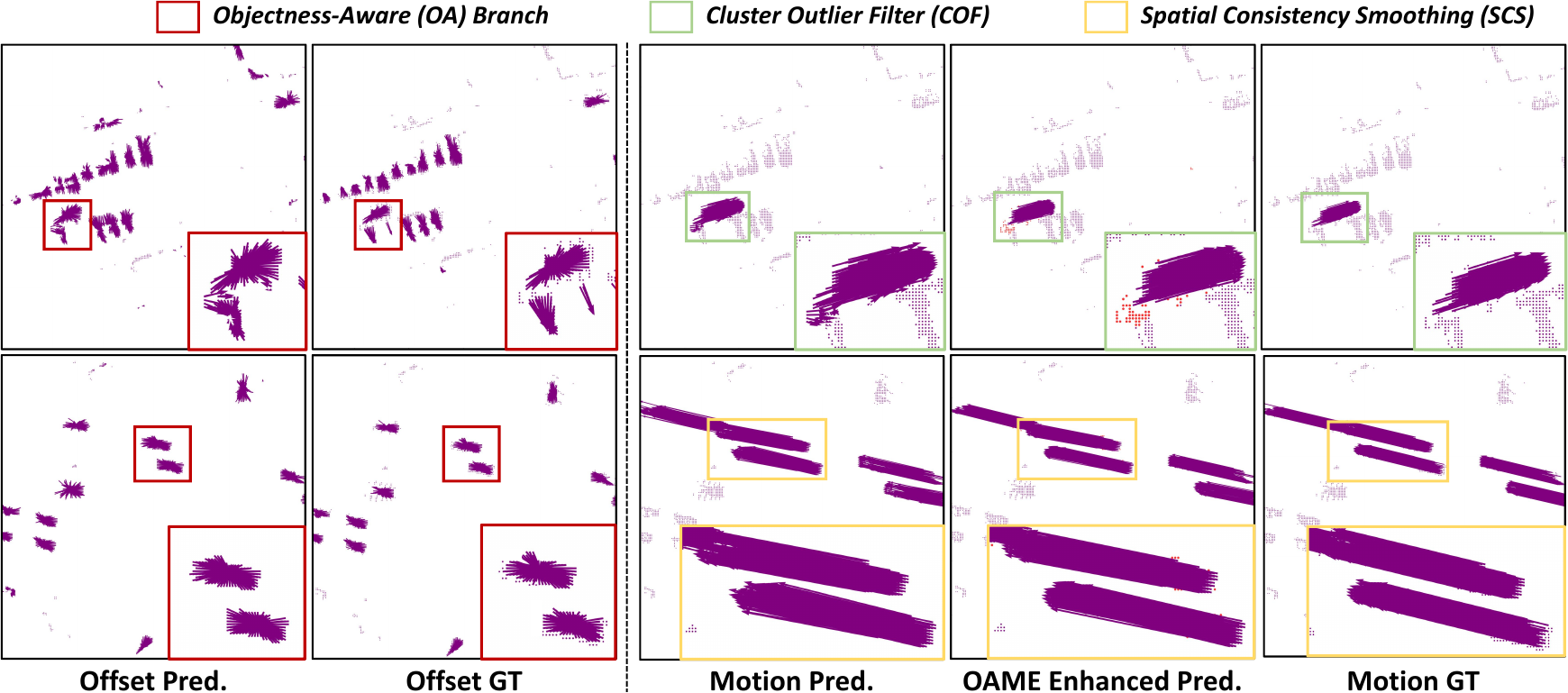}
\caption{\textbf{Visual evaluation of key components.} Red boxes indicate centroid shifts from the Objectness-Aware (OA) branch. Green boxes illustrate the effect of the Cluster Outlier Filter (COF) in suppressing noisy predictions. Yellow boxes demonstrate how the Spatial Consistency Smoothing (SCS) module enhances prediction coherence. OAME means the Objectness-Aided Motion Enhancement module, which consists of COF and SCS components. For clarity, predicted background points are omitted from the visualization.}
\label{fig:visualization_offset}
\vspace{-0.0cm}
\end{figure*}

\textbf{Objectness-aware motion prediction (OA).} We incorporate an objectness-aware offset branch into MotionNet as the objectness-aware motion prediction network, named OAMNet. We compare OAMNet with MotionNet to evaluate the effectiveness of objectness-aware branch. First, group \romannumeral1\ and \romannumeral2\ are synthetic-only approaches without knowledge translation. OAMNet in group \romannumeral2\ effectively reduces the mean error of static cells by $40.7\%$ compared with MotionNet in group \romannumeral1, which demonstrates that the acquired objectness knowledge can complement the motion regression and suppress jitters of background points. Besides, simply combining MotionNet with motion knowledge translation leads to noisy pseudo labels and performance decline (\romannumeral3\ v.s. \romannumeral1). In contrast, OAMNet shows strong robustness to domain shift and offers stable improvement (\romannumeral4\ v.s. \romannumeral2). This consistent improvement in generalization across domains demonstrates that the objectness prior is not only robust to domain distribution shifts but also effectively enhances motion prediction through implicit integration.

\textbf{Objectness-aided motion enhancement (OAME).} 
Though combining OAMNet with the motion knowledge translation framework obtains basic knowledge transfer, the noisy pseudo labels still contribute to unstable translation.
Thus, we propose OAME with COF and SCS to filter out outliers and improve the consistency of pseudo labels. Cooperating with COF in group \romannumeral5, the performance gap is reduced by $58.3\%$ at the static speed compared with group \romannumeral4, which proves that COF effectively filters out noisy jitters in static backgrounds. Based on the constraints of rigid objects from SCS, group \romannumeral7\ achieves the best performance and reduces the mean error at fast speed by $20.2\%$. However, when applying SCS without COF in group \romannumeral6, the spatial clusters contain numerous noisy outliers, disturbing the pseudo label smoothing and leading to inferior results. 
This also demonstrates the effectiveness of COF for filtering outliers and accurate clustering. 

\begin{figure*}[!t]
\centering
\includegraphics[width=0.9\linewidth]{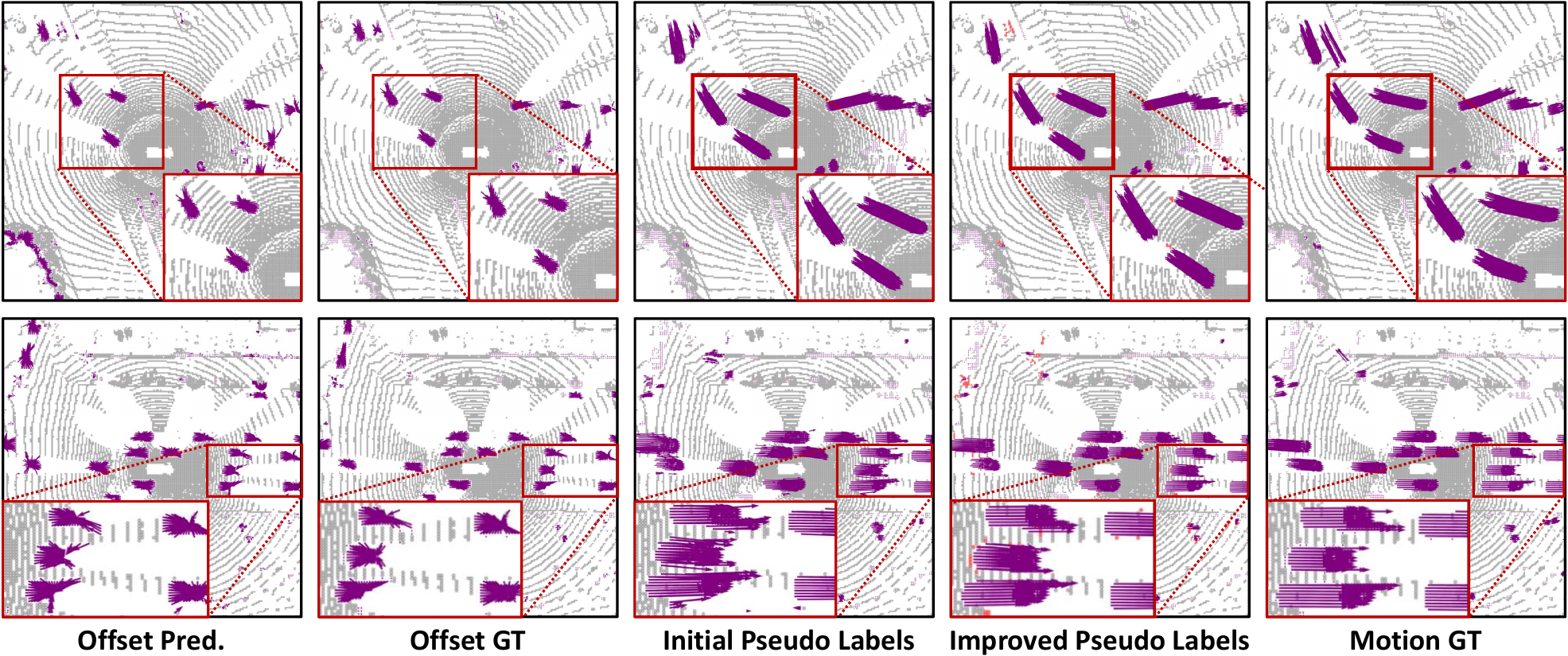}
\caption{\textbf{Visual comparisons of pseudo labels on Waymo training set.} Our approach effectively identifies and removes noisy labels (shown in orange), producing higher-quality results that closely approximate the ground truth. Key improvements are highlighted with red boxes.}
\label{fig:visualization}
\vspace{-0.2cm}
\end{figure*}

We also visualize the predicted center offsets and the behavior of the OAME module in Fig.~\ref{fig:visualization_offset}. Despite some noise in offset predictions under domain shift (e.g., first row), the majority of vectors point consistently toward object centers, validating the transferability of objectness knowledge. Crucially, our OAME module incorporates a dual-path consistency mechanism: only points where both the offset and motion branches agree on cluster assignment are retained. This geometric consensus acts as a built-in filter, i.e. COF module, suppressing false positives that may arise in one modality but are not corroborated by the other. This redundancy enhances robustness during synthetic-to-real transfer. Moreover, the SCS module further improves motion consistency within objects using the cleaned predictions from COF. As further illustrated in Fig.~\ref{fig:visualization}, combining offsets and motions, the OAME module filters out the noisy motion pseudo labels and produces high-quality pseudo labels.

\subsubsection{Comparison of different strategies in COF} 
Cluster outlier filtering is a strict two-stage filter, where a cell with either inaccurate motion or offset will be recognized as an outlier. As shown in Table~\ref{table:ablation-COF}, we loosen the constraints by employing the one-stage filter, where group cells in the current frames. As the cells with inaccurate motion prediction are reserved, the noisy pseudo labels result in inferior performance.  Besides, we observe that a large $\mathcal{T}_{N}$ filters out more noisy jitters on static cells but also excludes some correct predictions of small objects. Thus, we only filter out the clusters with a single cell ($\mathcal{T}_{N}=1$) as outliers for robust learning. 

\begin{table}[t]
\centering
\caption{Evaluation of different strategies of COF.}
\label{table:ablation-COF}
\begin{tabular*}{0.9\linewidth}{@{}@{\extracolsep{\fill}}c|ccc@{}}
\toprule
Strategy&Static&Slow&Fast\\
\midrule
One-stage &0.064&0.393 &1.597 \\
Two-stage & 0.048 & 0.385 & 1.545 \\
\midrule
$\mathcal{T}_{N} = 3$ &0.021&0.376 &2.035 \\
$\mathcal{T}_{N} = 1$ & 0.048 & 0.385 & 1.545 \\
$\mathcal{T}_{N} = 0$ & 0.068 & 0.387 & 2.174 \\
\bottomrule
\end{tabular*}
\vspace{-0.4cm}
\end{table}

\begin{table}[t]
\centering
\caption{Evaluation of different strategies of SCS.}
\label{table:ablation-SCS}
\begin{tabular*}{0.9\linewidth}{@{}@{\extracolsep{\fill}}c|cc|ccc@{}}
\toprule
Strategy&Smoothing&$\mathcal{T}_{CV}$&Static&Slow&Fast\\
\midrule
1 &\ding{55}&$\infty$&0.035&0.410 &1.936 \\
2&\checkmark&$\infty$&0.043&0.394 &1.637 \\
3&\ding{55}&10&0.039&0.366 &1.941 \\
\midrule
4&\checkmark&5& 0.041 & 0.407 & 1.677 \\
5&\checkmark&10& 0.048 & 0.385 & 1.545 \\
6&\checkmark&20 & 0.044 & 0.411 & 1.644 \\
\bottomrule
\end{tabular*}
\vspace{-0.4cm}
\end{table}

\subsubsection{Comparison of different strategies in SCS} 
After filtering out most of the unreliable cells, we apply spatial consistency smoothing to constrain pseudo labels with rigid spatial clusters. Besides, a selection operation is also included to exclude clusters with CVs larger than $\mathcal{T}_{CV}$. Setting $\mathcal{T}_{CV}$ to infinity ($\infty$) implies that this operation will not take effect. As shown in Table~\ref{table:ablation-SCS}, the motion smoothing in Eqn.~\ref{eq:cluster_mean} guarantees motion consistency within clusters and narrows the performance gap with the Oracle model by $31.7\%$ (strategy 2 v.s. strategy 1). The selection procedure improves the performance at the slow level (strategy 3 v.s. strategy 1).  We also conduct an ablation on thresholds $\mathcal{T}_{CV}$ (strategy 4-6). The lower $\mathcal{T}_{CV}$ is, the more noisy pseudo labels are excluded but also the fewer clusters are reserved. As we pay more attention to fast objects in autonomous driving systems, $\mathcal{T}_{CV}$ is determined as $10$ in our implementation. 

\subsubsection{Ablation on smoothing factor}
After every step $\tau$, the weight of the teacher model $\Phi^t$ is updated by the EMA weights of the student model $\Phi^s$ as formulated in Eq.\ref{eq:ema}, which can smooth the model noise and improve the performance of $\Phi^t$. 
To investigate the influence of the smoothing factor $\alpha$, we conduct an ablation study with different values, and the results are summarized in Table~\ref{table:smoothing_factor}. When $\alpha=1.0$, the teacher model remains static and does not incorporate any new knowledge from the student. In this case, the model fails to adapt to the real-world domain, preventing effective knowledge transfer from synthetic source data and resulting in significantly degraded performance. On the other hand, setting $\alpha=0.0$ (i.e., directly using the student as the teacher at every step) leads to highly unstable pseudo-labeling, as the teacher becomes overly sensitive to transient errors or fluctuations in the predictions of the student. This instability severely hinders training convergence and degrades overall accuracy.

These results clearly demonstrate that the EMA mechanism is not merely a smoothing trick, but a crucial component for balancing stability and adaptability. It enables the teacher to gradually absorb useful knowledge from the student while filtering out harmful noise, thus facilitating robust domain adaptation and reliable self-training. Based on this analysis, we set $\alpha=0.999$ to achieve optimal performance, ensuring both effective learning dynamics and label consistency.

\begin{table}[!t]
\small
\begin{center}
\caption{Comparison of models trained with different smoothing factors.}
\label{table:smoothing_factor}
\begin{tabular*}{0.9\linewidth}{@{}@{\extracolsep{\fill}}c|ccc@{}}
\toprule
$\alpha$ &Static&Slow&Fast\\
\midrule
0&\textbf{0.020}&0.474&2.229 \\
0.99&0.024&	0.472&	1.886\\
0.999&0.048&\textbf{0.385} &\textbf{1.545} \\
0.9999&0.067& 0.433 & 2.046 \\
1.0&0.099&0.433&2.045 \\
\bottomrule
\end{tabular*}
\end{center}
\vspace{-0.4cm}
\end{table}

\subsubsection{Ablation on Data augmentation}
For robust learning, we introduce augmentation for the input BEV maps of the student model upon the input sequences of the student model. As demonstrated in Table~\ref{table:augmentation}, training with augmentation significantly boosts performance, particularly benefiting the prediction accuracy for static cells. By introducing transformations such as rotations and flips, the model learns to maintain its predictive accuracy under different conditions, thereby enhancing its reliability and stability across various scenarios. This improvement underscores the effectiveness of our augmentation strategy in fostering a more robust synthetic-to-real knowledge translation.
\begin{table}[!t]
\small
\begin{center}
\caption{Ablation on data augmentation.}
\label{table:augmentation}
\begin{tabular*}{0.9\linewidth}{@{}@{\extracolsep{\fill}}c|ccc@{}}
\toprule
 &Static&Slow&Fast\\
\midrule
w Aug.&0.048&0.385 &1.545 \\
w/o Aug.&0.118&	0.402&	1.571\\
\bottomrule
\end{tabular*}
\end{center}
\vspace{-0.4cm}
\end{table}

\section{Conclusion}{
In this work, our primary focus lies in addressing the challenges of SRMP. Through our proposed novel SR-Motion, overcoming issues associated with naive regression methods, we successfully transfer motion knowledge to real-world data and obtain a motion prediction model robust to domain shift. 
\textbf{Limitation.}
Following previous arts, our method operates on a BEV map, losing the fine-grained geometry needed to model non-rigid motion. Consequently, it assumes objects are rigid and has limitations when applied to deformable agents.
}
\section*{Acknowledgment}
This research is supported by the Agency for Science, Technology and Research (A*STAR) under its MTC Programmatic Funds (Grant No. M23L7b0021). This research is also supported by the MoE AcRF Tier 2 grant (MOE-T2EP20223-0001).


\section{References Section}
\bibliographystyle{IEEEtran}
\bibliography{mybib}


\section{Biography Section}

\begin{IEEEbiography}[{\includegraphics[width=1in,height=1.25in,clip,keepaspectratio]{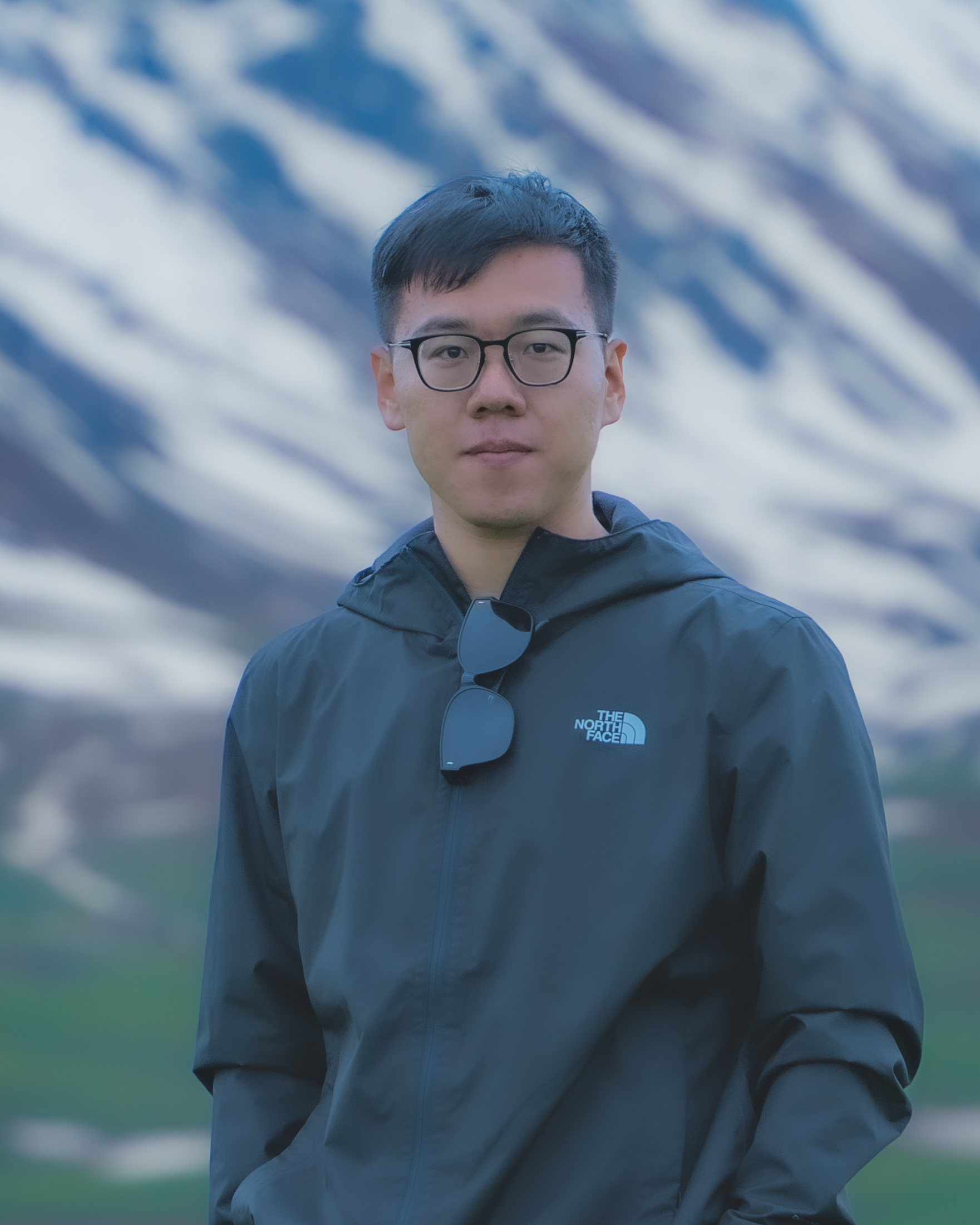}}]{Yizheng Wu} received the B.S. and M.S. degrees from Huazhong University of Science and Technology, Wuhan, China. He was a research assistant with the S-Lab, Nanyang Technological University, Singapore, from February 2023 to February 2024.
His reaserch interests include vision and language and instance segmentation.\end{IEEEbiography}\vspace{-0.0cm}

\begin{IEEEbiography}[{\includegraphics[width=1in,height=1.25in,clip,keepaspectratio]{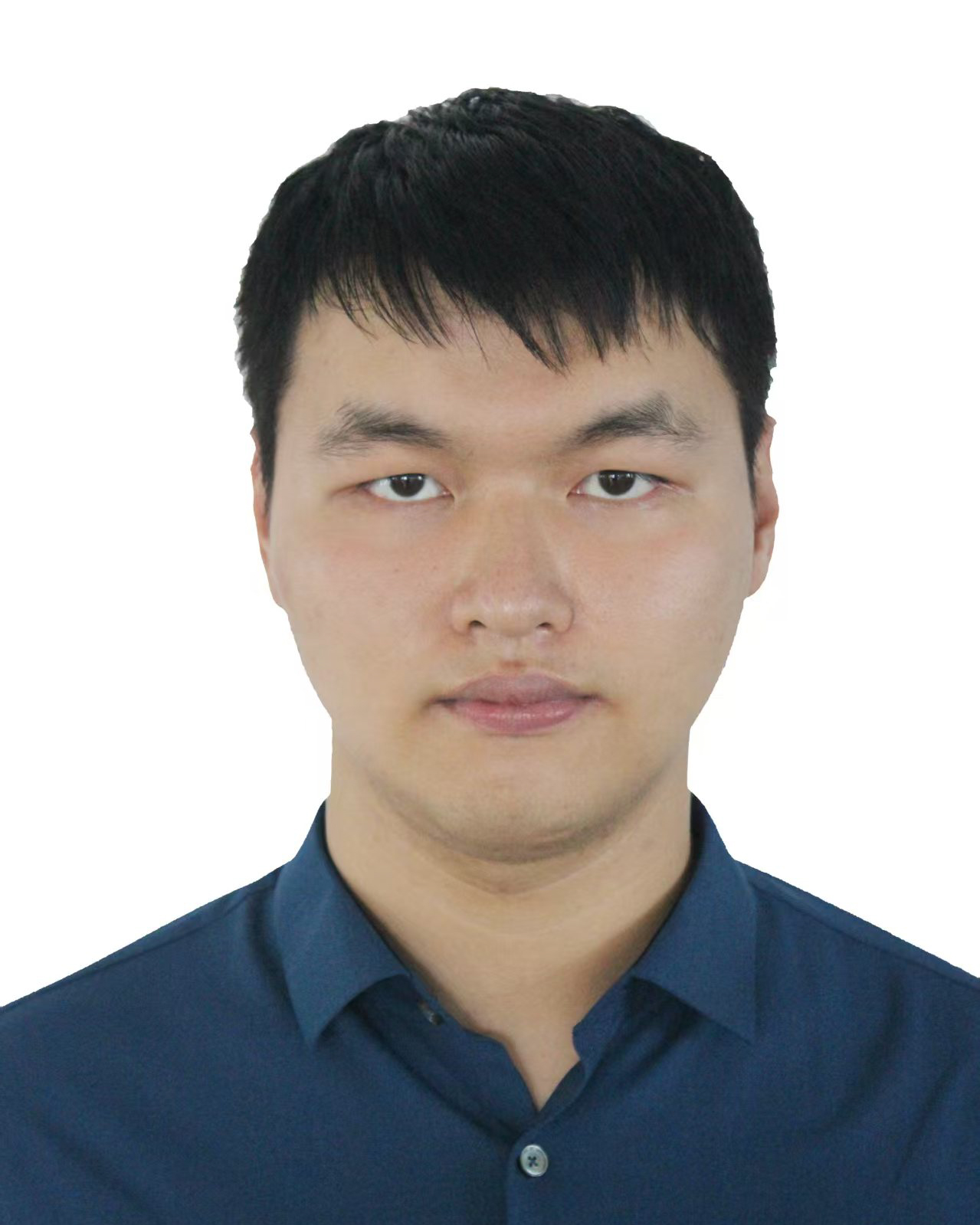}}]{Hongwei Fan} is now a first-year PhD student in the School of Computer Science, Peking University, advised by associate professor Hao Dong. His research interests include robotics, embodied AI, and 3D vision.\end{IEEEbiography}\vspace{-0.0cm}

\begin{IEEEbiography}[{\includegraphics[width=1in,height=1.25in,clip,keepaspectratio]{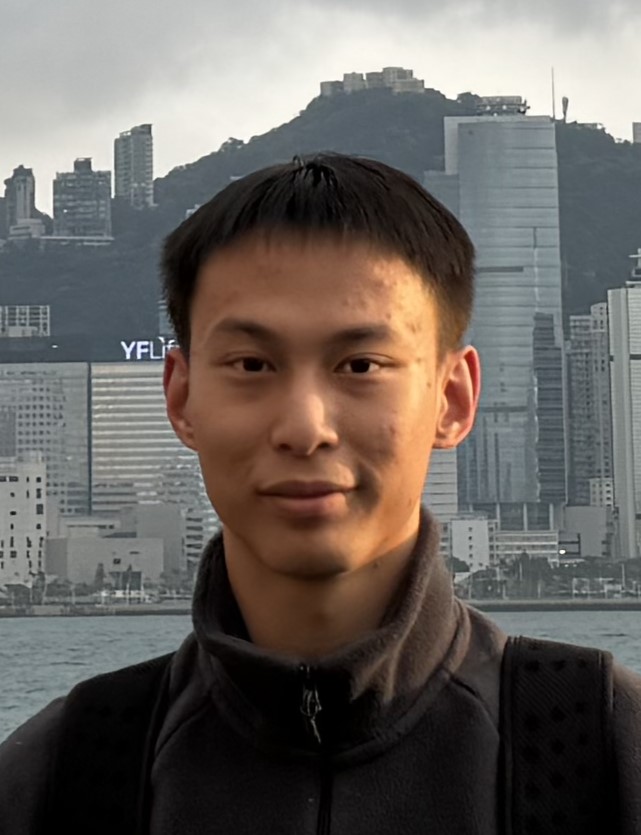}}]{Kewei Wang}
received the B.S. and M.S. degrees from Huazhong University of Science and Technology, Wuhan, China. He was a research assistant with the S-Lab, Nanyang Technological University, Singapore, from February 2023 to February 2024.
He has published papers in the area of motion prediction and semi-supervised learning.\end{IEEEbiography}\vspace{-0.0cm}

\begin{IEEEbiography}[{\includegraphics[width=1in,height=1.25in,clip,keepaspectratio]{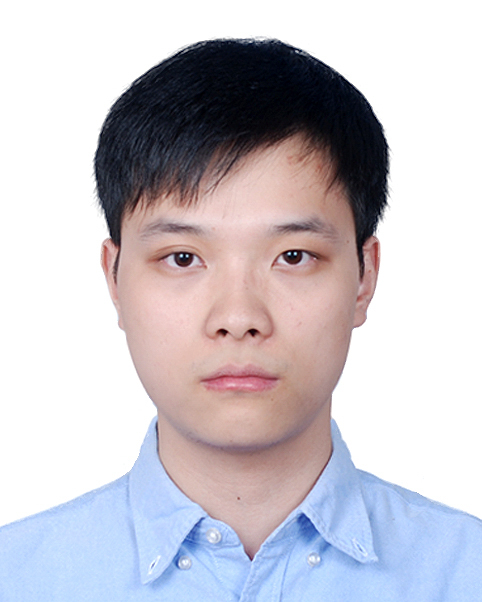}}]{Ruibo Li}
received the PhD degree from the School of Computer Science and Engineering at Nanyang Technological University in Singapore. He is currently a research fellow at Nanyang Technological University, Singapore. His research interests are in computer vision and machine learning.\end{IEEEbiography}\vspace{-0.0cm}

\begin{IEEEbiography}[{\includegraphics[width=1in,height=1.25in,clip,keepaspectratio]{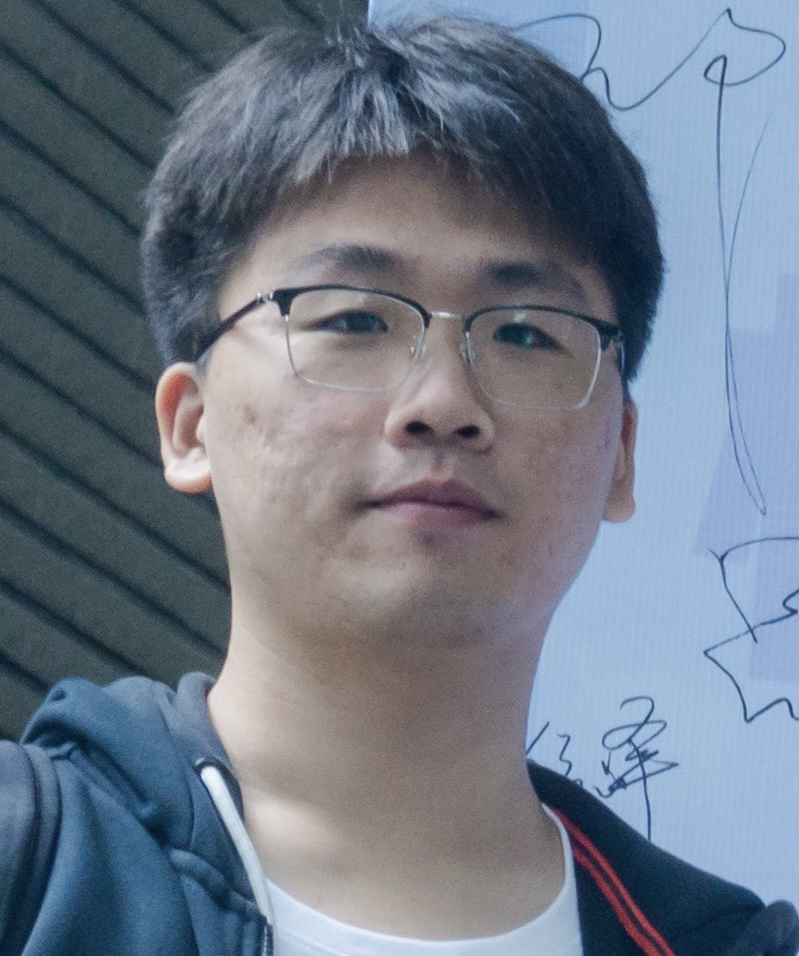}}]{Xingyi Li}
received the B.S. and M.S. degrees from the Huazhong University of Science and Technology, Wuhan, China. He was a research assistant with the S-Lab, Nanyang Technological University, Singapore, from February 2023 to February 2024.
He has published papers in the area of neural rendering and novel view synthesis.\end{IEEEbiography}\vspace{-0.0cm}

\begin{IEEEbiography}[{\includegraphics[width=1in,height=1.25in,clip,keepaspectratio]{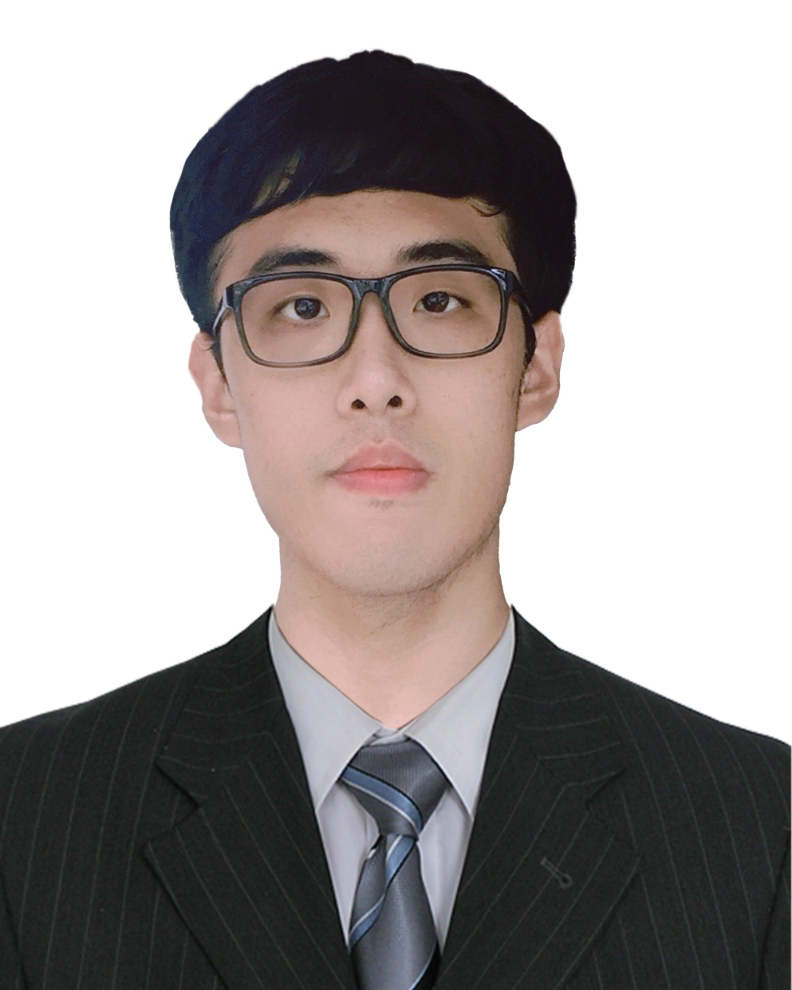}}]{Xiao Song} is now a senior researcher in the Department of Autonomous Driving at SenseTime Group Limited. His research interests include computer vision, deep learning, 3D perception and autonomous driving.\end{IEEEbiography}\vspace{-0.0cm}

\begin{IEEEbiography}[{\includegraphics[width=1in,height=1.25in,clip,keepaspectratio]{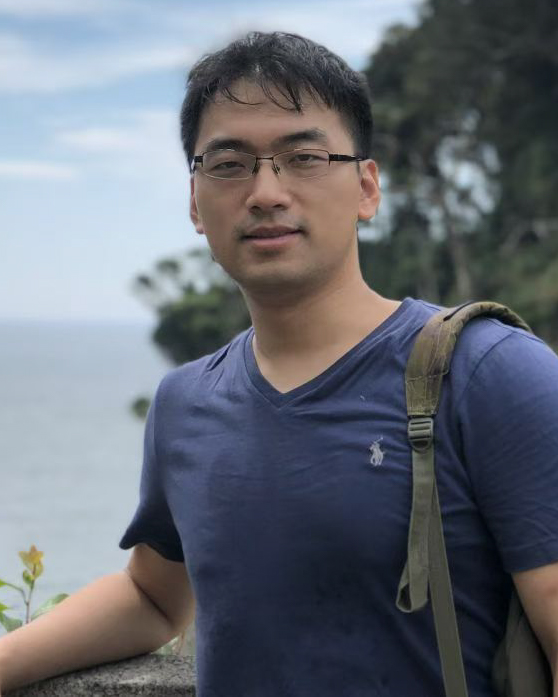}}]{Zhe Wang} received his B.S. degree in Optical Engineering of Zhejiang University in 2012, and the Ph.D. degree in the Department of Electronic Engineering at The Chinese University of Hong Kong. He is currently a director in SenseTime. His research interests include computer vision and deep learning.\end{IEEEbiography}\vspace{-0.0cm}

\begin{IEEEbiography}[{\includegraphics[width=1in,height=1.25in,clip,keepaspectratio]{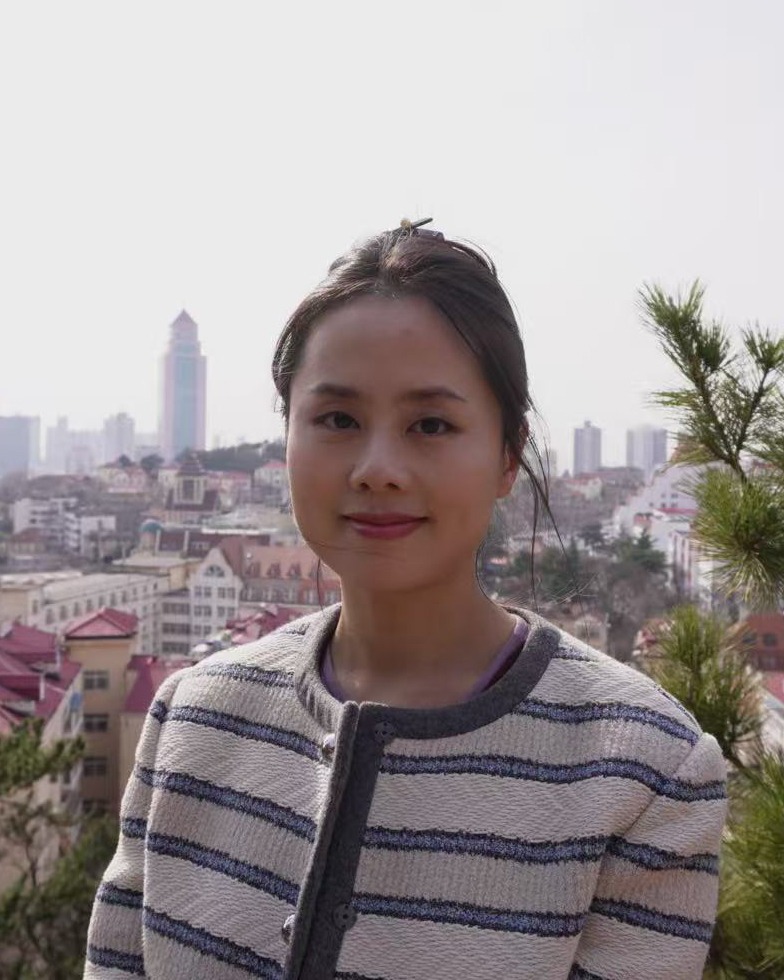}}]{Chenjing Ding} received her B.S. and M.S. degrees from Southeast University, and have been working at SenseTime ever since graduation. Her research interests since then include 3D reconstruction, video generation, and world models.\end{IEEEbiography}\vspace{-0.0cm}

\begin{IEEEbiography}[{\includegraphics[width=1in,height=1.25in,clip,keepaspectratio]{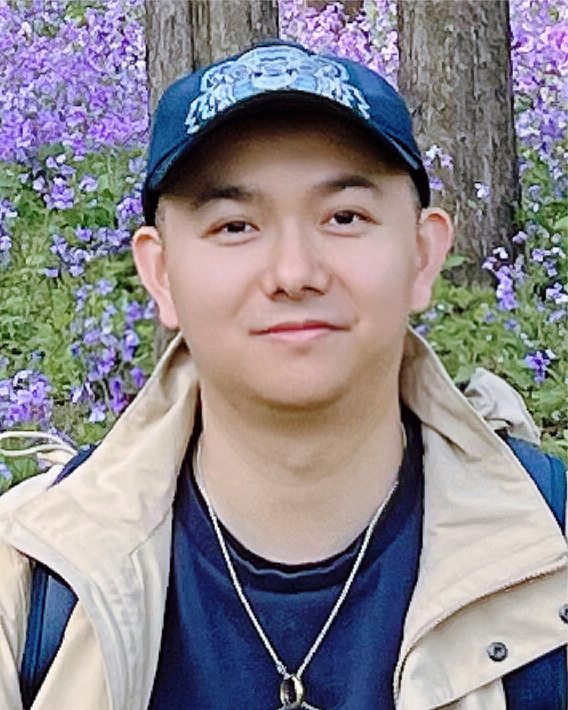}}]{Dongliang Wang} is a researcher at SenseTime, specializing in computer vision with a particular focus on video understanding, 3D content generation, and autonomous driving. He has contributed to the field through the publication of several academic papers on these topics. He completed his education at Xi'an Jiaotong University with distinction. And he was PhD candidate in the MSR(Asia)-XJTU jointed PhD project.\end{IEEEbiography}\vspace{-0.0cm}

\begin{IEEEbiography}[{\includegraphics[width=1in,height=1.25in,clip,keepaspectratio]{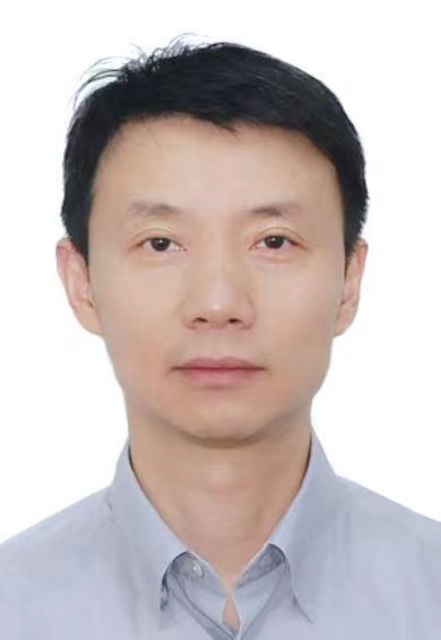}}]{Zhiguo Cao (Member, IEEE)} received the BS and MS degrees in communication and information system from the University of Electronic Science and Technology of China and the PhD degree in pattern recognition and intelligent system from the Huazhong University of Science and Technology. He is currently a professor with the School of Artificial Intelligence and Automation, Huazhong University of Science and Technology. His research interests include computational photography, monocular depth estimation, 3D video processing, motion detection, and human action analysis.\end{IEEEbiography}

\begin{IEEEbiography}[{\includegraphics[width=1in,height=1.25in,clip,keepaspectratio]{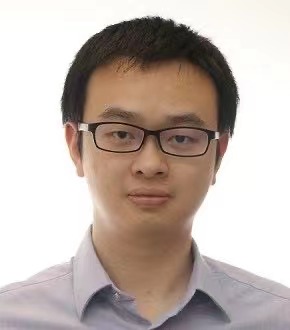}}]{Guosheng Lin} is an Associate Professor at the School of Computer Science and Engineering, Nanyang Technological University, Singapore. He received his PhD degree from The University of Adelaide in 2014. His research interests are generally in computer vision and machine learning including scene understanding, 3D vision and generative learning.\end{IEEEbiography}\vspace{-0.0cm}

\vfill

\end{document}